%% file: neuralvdb_arxiv.tex
\documentclass[acmtog,nonacm]{acmart}

\AtBeginDocument{%
  }

\begin{document}

\title{NeuralVDB: High-resolution Sparse Volume Representation using Hierarchical Neural Networks}
\author{Doyub Kim}
\affiliation{%
  \institution{NVIDIA}
  \country{USA}}
\email{doyubkim@nvidia.com}

\author{Minjae Lee}
\affiliation{%
  \institution{NVIDIA}
  \country{USA}}
\email{minjael@nvidia.com}

\author{Ken Museth}
\affiliation{%
  \institution{NVIDIA}
  \country{USA}}
\email{kmuseth@nvidia.com}

\renewcommand{\shortauthors}{Kim, Lee, and Museth}

\newcommand*{\ken}{}
\newcommand*{\minjae}{}
\newcommand*{\doyub}{}

\newcommand*{\kenmr}{}
\newcommand*{\minjaemr}{}
\newcommand*{\doyubmr}{}

\begin{abstract}
    \input{0_abstract.tex}
\end{abstract}



\begin{teaserfigure}
    \includegraphics[width=\textwidth]{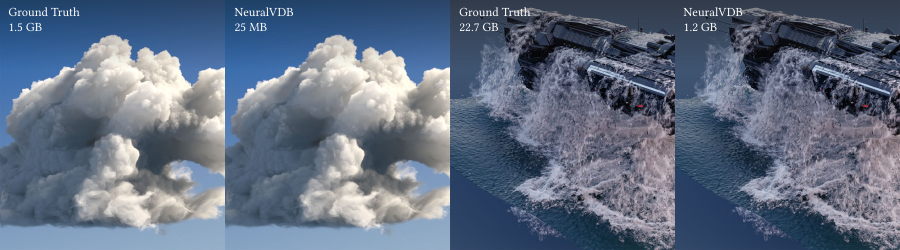}
    \caption{Application of NeuralVDB to the Disney Cloud dataset~\cite{disneyclouds} (left) and a time-series of narrow-band level sets of an animated water surface generated from a high-resolution simulation of a space ship breaching rough sea (right). The file size of the Disney Cloud, represented by the industry standard {\bf OpenVDB} encoded with 16-bit quantization and Blosc compression, is $1.5GB$. However, the corresponding {\bf NeuralVDB} file only has a footprint of $25MB$, resulting in a reduction by a factor of $\mathbf{60}$. For the space ship breaching simulation, the \minjae{accumulated} file sizes for the entire sequence of the water surface, using OpenVDB with the same compression (16-bit and Blosc), is $22.7GB$, whereas the NeuralVDB representations only have a total footprint of $1.2GB$, corresponding to a reduction by a factor of $\mathbf{18}$.}
    \label{fig:teaser}
\end{teaserfigure}

\maketitle

\input{body}

\bibliographystyle{ACM-Reference-Format}
\bibliography{references}

\appendix

\input{9_appendix}

\end{document}

%% file: 0_abstract.tex
 
We introduce NeuralVDB, which improves on an existing industry standard for efficient storage of sparse volumetric data, denoted VDB~\cite{museth2013vdb}, by leveraging recent advancements in machine learning. Our novel hybrid data structure can reduce the memory footprints of VDB volumes by orders of magnitude, while maintaining its flexibility and only incurring small (user-controlled) compression errors. Specifically, NeuralVDB replaces the lower nodes of a shallow and wide VDB tree structure with multiple hierarchical neural networks that separately encode topology and value information by means of neural classifiers and regressors respectively. This approach is proven to maximize the compression ratio while maintaining the spatial adaptivity offered by the higher-level VDB data structure. For sparse signed distance fields and density volumes, we have observed compression ratios on the order of $10\times$ to more than $100\times$ from already compressed VDB inputs, with little to no visual artifacts. \doyubmr{Furthermore, NeuralVDB \minjaemr{is shown} to offer more effective compression performance compared to other neural representations such as Neural Geometric Level of Detail~\cite{takikawa2021neural}, Variable Bitrate Neural Fields~\cite{takikawa2022variable}, and Instant Neural Graphics Primitives~\cite{mueller2022instant}}\kenmr{. Finally, we} demonstrate how warm-starting from previous frames can accelerate training, i.e.,~compression, of animated volumes as well as improve temporal coherency of model inference, i.e.,~decompression.

%% file: body.tex
\section{Introduction}
\label{sec:introduction}
\input{1_introduction.tex}

\section{Related work}
\label{sec:related-work}
\input{2_related_work.tex}

\section{Method}
\label{sec:method}
\input{3_method.tex}

\section{Results}
\label{sec:results}
\input{4_results.tex}

\section{Discussion}
\label{sec:discussion}
\input{5_discussion.tex}

\section{Acknowledgements}
\label{sec:acknowledgements}
\input{8_acknowledgements.tex}

%% file: 1_introduction.tex
Sparse volumetric data are ubiquitous in many fields \minjae{including} scientific computing and visualization, medical imaging, industrial design, rocket science, computer graphics, visual effects, robotics, and more recently machine learning applications. As such it should come as no surprise that several compact data structures have been proposed over the years for efficient representations of sparse volumes. One of these sparse data structures has gained widespread adoption in especially the entertainment industry, namely OpenVDB, and is showing signs of increased adoption in several other fields~\cite{achilles2016medical,boddeti2020manufacturing,vizzo2022robotics}.

OpenVDB is based on the unique hierarchical tree data structure introduced by ~\cite{museth2013vdb}. At the core it is a shallow (typically four-level) tree with high but varying fanout factors (e.g.,~$32^3\rightarrow16^3\rightarrow8^3$---number of nodes per level from top to bottom), and the ability to efficiently look up values through fast bottom-up, \minjae{vs.} slower top-down, node access patterns. While its initial open source implementation, OpenVDB, was limited to CPUs, a read-only GPU variant, dubbed NanoVDB, was recently developed~\cite{museth2021nanovdb} and added to the \kenmr{open source} library. However, VDB is obviously not a silver bullet, and fundamentally suffers from the same limitations as other lossless volumetric data \minjae{structures:} the memory footprint is never smaller than that incurred by the sparse non-constant voxel values, e.g.,~signed distance or density values. To a lesser extent the same is true for the topology-information of the sparse voxels, which are compactly encoded into \minjaemr{bitmasks} of the tree nodes in VDB. \doyubmr{To provide some context, the Disney Cloud is 1.5 GB with conventional data compression techniques and 16-bit quantization \minjaemr{(shown in Figure~\ref{fig:teaser}).} This size can easily explode into terabytes of data per simulation sequence or high-resolution volumetric \minjaemr{scenes.} These data sets are frequently shared between data consumers and/or cloud storage, where \kenmr{both data storage and transactions are typically costly}. While many scenarios require raw, lossless data, \minjaemr{other workflows can tolerate some degree of lossy compression in exchange for a lighter data footprint, akin to using JPEG images in place of raw images.}} This raises the question, are there more compact, \minjae{possibly lossy,} representations for the topology and value information encoded into a VDB structure, that \minjae{maintain} many of the advantages of the proven VDB tree structure? 

We will spend the remainder of this paper demonstrating, that under the same assumptions as NanoVDB, i.e.,~static topology and values, this is indeed the case, resulting in a new hybrid data structure, which we have dubbed NeuralVDB.

The key to unlocking the promise of NeuralVDB is, as the name indicates, neural networks. Recently neural scene representations have gained a lot of attention from the research community, especially around implicit geometry~\cite{park2019deepsdf,mescheder2019occupancy,michalkiewicz2019implicit,liu2020neural} or radiance fields~\cite{mildenhall2020nerf,yu2020pixelnerf}. Essentially, the neural representation encodes the field function that maps multi-dimensional input (such as positional coordinates or directions) to a field value (such as SDF, occupancy, density, or radiance) using neural networks. Thanks to the flexibility and differentiability of neural networks, this new approach opened up a variety of applications, including novel view reconstruction~\cite{mildenhall2020nerf}, compression~\cite{davies2020effectiveness,li2022high,takikawa2022variable}, adaptive resolution~\cite{takikawa2021neural}, etc. \doyubmr{Nonetheless, as we will illustrate in Section~\ref{sec:comparison} through additional comparisons with established neural scene representation techniques, relying solely on a neural approach falls short in delivering a model that balances both high quality and compact size. \minjaemr{By hybridizing a state-of-the-art data structure with a neural representation, NeuralVDB surpasses other methods in both qualitative and quantitative measures.}}

We propose a new approach to memory efficient representations of static sparse volumes that combines the best of two worlds: neural scene representations have demonstrated that neural networks can achieve impressive compression of 3D data, and VDB offers an efficient hierarchical partitioning of sparse 3D data. \kenmr{This} combination allows a VDB tree to focus on coarse upper node level topology information, while multiple neural networks compactly encode fine-grain topology and value information at the voxel and lower tree levels. This also applies to animated volumes, even maintaining temporal coherency and improving performance with our novel temporal encoding feature.  

\ken{
We outline the goals, non-goals, and constraints of NeuralVDB as follows:
\begin{itemize}
    \item The overarching {\bf goal} of NeuralVDB is to significantly reduce both the off-line, e.g., file, and on-line, e.g., memory, footprints of sparse volumetric data represented with the VDB data structure.  We achieve this goal by means of compact neural representations of both the spatial occupancy, i.e., topology, and the values of the sparse volumes.
    \item A {\bf non-goal} of NeuralVDB is to improve the speed of volume rendering. That is, we are willing to sacrifice rendering speeds for the sake of reducing the file or memory footprints. While we make efforts to minimize this performance trade-off, and even offer two versions of NeuralVDB with different ratios of compression to access-performance, we emphasize that the objective of this paper is not to propose a faster data structure for volume rendering.
    \item An important design {\bf constraint} of NeuralVDB is to preserve \minjae{information represented in the input VDB volumes} as much as possible, as well as \minjae{to} maintain compatibility with existing VDB pipelines. That is, we reuse the VDB tree structure and its API as such as possible, use lossless compression of spatial occupancy, i.e., topology information, and adaptive lossy compression for the values of the sparse volumes.
\end{itemize}
}

More precisely we summarize our contributions as follows:

\paragraph{Memory Efficiency} The main focus of NeuralVDB is data compression, both out-of-core and in-memory. In contrast, OpenVDB only provides out-of-core compression, like Blosc and Zlib~\cite{gailly2004zlib}. In-core representations of OpenVDB apply no compression to the sparse values, and only per-node \minjaemr{bitmask} compression of the topology, i.e.,~sparse coordinates. While NanoVDB improves on OpenVDB by offering in-core variable \minjaemr{bitrate} quantization of the sparse values, the compression ratio of NanoVDB rarely exceeds $6\times$, when low quantization noise is desired. Conversely, for in-core representations NeuralVDB typically offers an order of magnitude higher compression ratio than NanoVDB, and two orders of magnitude higher compression ratio than OpenVDB. However, neither data-agnostic compression techniques like Zlib nor bit-quantization leverage feature level similarities of sparse voxels.

Neural networks, on the other hand, can be designed to discover such hidden features and can infer values without reconstructing the entire data set. NeuralVDB exploits
such characteristics of neural networks to effectively compress volumetric data while simultaneously supporting random access.

\paragraph{Compatibility} NeuralVDB is designed to be compatible with existing VDB pipelines. Specifically, NeuralVDB representations can readily be encoded from VDB data and decoded back into VDB representations, with small often invisible reconstruction errors. Borrowing standard terminology from machine learning we refer to these steps as training and inference, respectively. While NeuralVDB is designed to encode topology information exactly, values are encoded with a lossy compressor whose key objective is to retain as much information as possible during the training. For instance, a NeuralVDB structure shares the same higher level tree structure with standard VDB. The hierarchical network, which replaces the lower level structure is also designed to reconstruct the original VDB tree levels. As such, NeuralVDB supports both out-of-core and in-core decompression, which can be utilized respectively as an offline compression codec or alternatively for online applications like rendering that require direct in-memory access. 

The remainder of this paper is organized as follows: in Section~\ref{sec:related-work} we review related work, followed by a brief summary of the key features of VDB and the framework supporting NeuralVDB in Section~\ref{sec:method}.
Finally, we validate our performance claims of NeuralVDB in Section~\ref{sec:results} and conclude with a discussion of limitations and future work in Section~\ref{sec:discussion}.

%% file: 2_related_work.tex
In this section, we review previous studies discussing efficient representation and computation of sparsely distributed volumetric data.

\subsection{Data Compression}

While there is a wide variety of algorithms for data compression, we shall limit our discussion to three subcategories that best highlight the difference between traditional compression techniques and the novel approach of NeuralVDB.

The first category of compression techniques includes \emph{data-agnostic} algorithms like Zlib~\cite{gailly2004zlib}. As mentioned in the previous section, these algorithms are great at compressing arbitrary data, but by design cannot exploit geometric structures or patterns present in the data. It can, however, be utilized to compress the last layer of our neural networks. For instance, similarly to OpenVDB, NeuralVDB uses Blosc~\cite{blosc} to compress the serialized buffer.

The second class of compression techniques is best described as \emph{application-specific} algorithms similar to JPEG~\cite{pennebaker1992jpeg} for images or MPEG~\cite{le1991mpeg} for videos. The extension of 2D JPEG algorithms to 3D can be a good candidate for volumetric data. However, it is not directly applicable to VDB, since JPEG is based on spectral analysis of 2D images (by means of discrete cosine transformations), which operates on dense domains, whereas VDB is inherently sparse in 3D. However, we have seen promise in recent studies \kenmr{that employ} neural networks for compression problems~\cite{ma2019image,kirchhoffer2021overview} or even combining conventional compression techniques with neural approaches~\cite{liu2018deepn} to exceed the compression performance of the original algorithm.
There are mesh based compression methods~\cite{pajarola2000compmesh,valette2004wavemesh,sattler2005companim}, which can only handle meshes \kenmr{as oppose to sparse} volumes.

Lastly, the third type of compression is \emph{statistical approaches} such as principal component analysis (PCA) or auto-encoders (AE). These techniques are based on learned models that are derived from training data. By transforming the input space into a reduced latent space, high dimensional input data can be represented with relatively small-sized vectors. In fact, some of the earlier studies on neural-implicit representation, such as DeepSDF~\cite{park2019deepsdf}, utilize AE to further compress the SDF volumes. This approach, however, requires the input space to be known and/or normalized into a known shape. NeuralVDB takes a different approach in that it deliberately "over-fits" to the input volume, i.e.,~memorizes the input as much as possible. This approach trades off statistical knowledge that could be learned from data with flexibility that can take arbitrary inputs.

\subsection{Sparse Grid}

While there is a large body of work on sparse data structures in computer graphics, we shall limit our discussion to sparse grids in the context of numerical simulation and rendering, which are the core target applications of NeuralVDB. 

One such key application is level set methods, which are essentially time-dependent truncated signed distance fields (SDF). These are efficiently implemented with narrow-band methods that track a deforming zero-crossing interface~\cite{peng1999pde}. Additional memory efficiently has been demonstrated with adaptive structures like octree grids~\cite{strain2001fast,losasso2004simulating,bargteil2006semi}, Dynamic Tubular Grids (DT-Grid, based on compressed-row-storage)~\cite{NielsenMuseth2006}, or tall-cell grids~\cite{irving2006efficient,chentanez2011real}.

More flexible data structures for generic simulation and data types include Hierarchical Run-length Encoding (HRLE) grid~\cite{houston2006hierarchical}, B+Grid (precursor to VDB)~\cite{DB+Grid}, VDB (open sourced as OpenVDB)~\cite{museth2013vdb}, Field3D (tiled dense grid)~\cite{field3d}, Sparse Paged Grid (SPGrid, inspired by VDB)~\cite{setaluri2014spgrid},  GVDB (loosely based on VDB)~\cite{hoetzlein2016gvdb}, KDSM (Kinematically Deforming Skinned Mesh)~\cite{lee2018kdsm,lee2019kdsm} and more recently NanoVDB (strictly based on VDB)~\cite{museth2021nanovdb}.

\subsection{Neural Representation}

The idea of utilizing neural networks to represent volumetric data is by no means novel. Examples include occupancy field~\cite{mescheder2019occupancy,peng2020convolutional}, implicit surface like SDF~\cite{michalkiewicz2019implicit,park2019deepsdf,mescheder2019occnet,chen2018impdec,tang2020deep,tang2018real}, and multi-dimensional data like radiance field~\cite{mildenhall2020nerf} are encoded using neural networks. Most of these studies utilize coordinate-based neural networks and feature mapping/encoding techniques such as SIREN~\cite{sitzmann2020implicit}, Fourier Feature Mapping~\cite{tancik2020fourfeat}, and Neural Hashgrid~\cite{mueller2022instant}. We refer readers to \cite{xie2022nf} for a general survey on neural fields.

\subsection{Hybrid Methods}

The desire for neural representations that are both memory efficient and allow for fast random queries, has led to the development of hybrid methods that combines neural networks and sparse data structures. Recent examples hereof are Neural Sparse Voxel Fields~\cite{liu2020neural}, Neural Geometric Level of Detail ~\cite{takikawa2021neural}, Baking NeRF~\cite{hedman2021baking}, and Adaptive Coordinate Networks~\cite{martel2021acorn}. Learning a tree data structure indexing was also presented in ~\cite{kraska2018case}.

NeuralVDB also falls into this category. The main difference between existing hybrid methods and NeuralVDB lies in the key design goals we mentioned earlier -- better memory efficiency and compatibility with VDB. While the previous hybrid approaches are memory-efficient compared to conventional neural representations, they are less efficient compared with the non-neural sparse grid structures. We carefully allocate and train parameters such that NeuralVDB can achieve high-fidelity reconstruction while consuming much less memory than compressed VDB. Also, NeuralVDB is compatible with existing VDB pipelines by design and can retain input (standard) VDB's original hierarchical structure with minimal error. Additionally, NeuralVDB is not limited to specific types of volumes such as occupancy, signed distance field, volume density, or even vector fields. Finally, NeuralVDB is an open framework that does not require a dedicated network architecture. Therefore, any purely neural or even hybrid methods can be used as a black box submodule of NeuralVDB.

%% file: 3_method.tex
This section will briefly outline the original VDB tree structure and explain how it is used to derive NeuralVDB, which combines explicit tree and implicit neural representations. More precisely, we demonstrate how different neural networks can be designed to separately encode topology and value information in NeuralVDB. We demonstrate how the decoder in NeuralVDB can be used for both offline/out-of-core and online/in-memory applications. Finally, we \kenmr{introduce a novel temporal warm-starter that encodes animated VDBs with improved training performance and temporal coherency of the reconstructed VDBs}.

\subsection{VDB}
\label{sec:standard-vdb}

\begin{figure*}[t]
    \centering
    \includegraphics[width=6in]{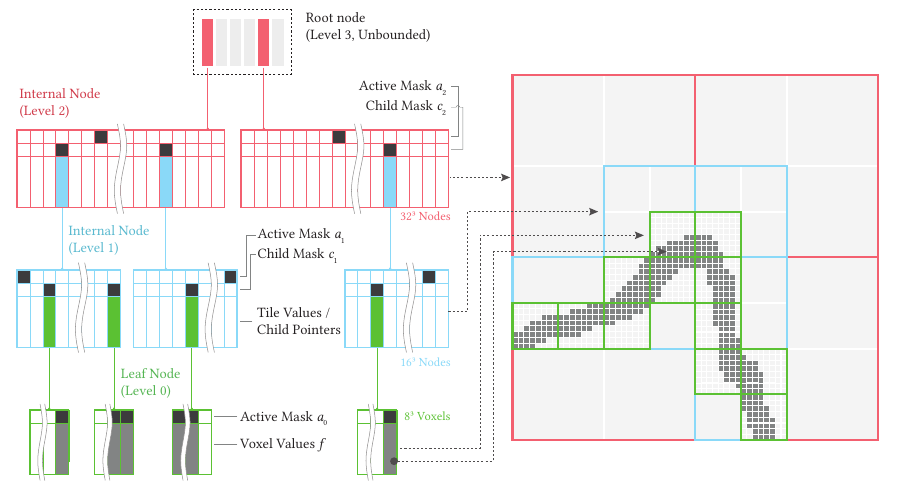}
    \caption{1D and 2D illustrations of VDB data structures. {\bf Left:} A 1D 4-level VDB tree hierarchy is shown with its various node structures and bitmasks. The top-most root node (level 3) holds an unbounded set of internal nodes (level 2), and the red/blue internal nodes encode tile values or child pointers using bitmasks ($a_l$ and $c_l$). The lower green leaf nodes store voxel values $f$ and their active masks $a_0$. {\bf Right:} 2D illustration of the hierarchical tree nodes that intersect the sparse (gray) pixels. The color schemes are shared between the 1D and 2D illustrations. \doyubmr{The number of nodes per level are indicated where the level 2 and 1 internal nodes have $32^3$ and $16^3$ children nodes and level 0 (leaf) nodes have $8^3$ voxels per node.}}
    \label{fig:vdb}
\end{figure*}

Let us briefly summarize the main characteristics of a VDB tree structure as well as its unique terminology. (For more details we refer the reader to the original paper~\cite{museth2013vdb}).

In a VDB tree structure, values are associated with all levels of the tree, and exist in a binary state referred to as respectively active or inactive values. Specifically, values at the leaf level, i.e., the smallest addressable (integer) coordinate space, are denoted  \textbf{voxels}, whereas values residing in the upper node levels are referred to as \textbf{tile values}, and cover larger coordinate domains. \ken{That is, tile values conceptually corresponding to uniform values assigned to all voxels subsumed by the node that the tile resides in, thus compactly representing constant regions of space}.  While the VDB tree structure, detailed in \cite{museth2013vdb}, can have arbitrarily many configurations, we will exclusively focus on the default configuration used in OpenVDB, which has proven useful for most practical applications of VDB.  This configuration uses four levels of a tree with a sparse unbounded root node followed by three levels of dense nodes of coordinate domains $4096^3$, $128^3$ and $8^3$. Thus, leaf nodes can be thought of as small dense grids of size $8\times8\times8$, arranged in a shallow tree of depth four with variable fanout factors ($n$ as in $n^3$, the number of nodes per level) of $32$ and $16$ respectively. We will refer to the leaf level as level 0, internal nodes as level 1 and 2, and the top-most root level as level 3.  Thus, a default VDB tree can be implemented as a hash table of dense child nodes of size $32^3$, each with dense child nodes of size $16^3$, each with dense child nodes of size $8^3$. Figure~\ref{fig:vdb} illustrates this tree structure in one and two spatial dimensions. Finally, note that all internal nodes (at level 1 and 2) have two bitmasks, denoted \textbf{active mask $a_l$} and \textbf{child mask $c_l$}, which respectively indicate if a tile value is active or whether it is connected to a child node. Conversely, leaf nodes only have an \textbf{active mask $a_l$} used to distinguished active vs inactive voxels.

Throughout this paper we will adopt the same notation for VDB tree configurations that was introduced in \cite{museth2013vdb}. Thus, the configuration outlined above, which is the default in OpenVDB, is denoted $[\textrm{Hash},5,4,3]$, where $\textrm{Hash}$ refers to the fact that the root node employs a sparse hash-table whereas the remaining tree levels are dense, i.e.,~fixed-size, with nodes logarithmic sizes $5,4,3$, corresponding to the dimensions $32^3,16^3,8^3$, which in turn covers the coordinate domains $4096^3$, $128^3$ and $8^3$. 
\ken{In the appendix we explain how VDB facilitates fast random access, and how NanoVDB offers GPU acceleration \cite{museth2021nanovdb}.}

\subsection{NeuralVDB}
\label{sec:vdb-with-nn}

\begin{figure*}[t]
    \centering
    \includegraphics[width=6in]{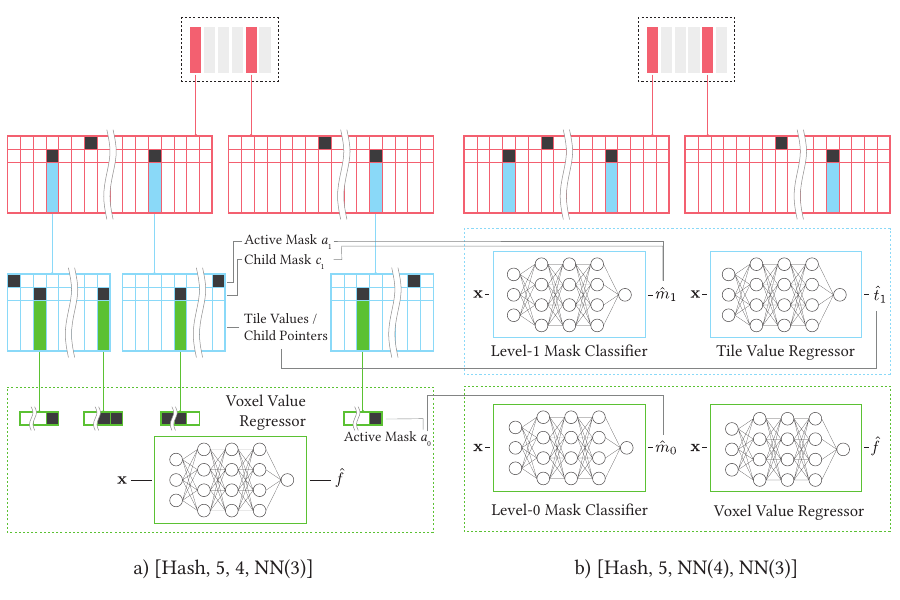}
    \caption{Illustration of two different NeuralVDB structures: a) a standard VDB tree with neural voxel values ($[\textrm{Hash},5,4,\textrm{NN}(3)]$ using the VDB tree notation), and b) a hybrid VDB/neural tree with neural representations of both nodes and their values ($[\textrm{Hash},5,\textrm{NN}(4),\textrm{NN}(3)]$ using the tree notation).}
    \label{fig:overview}
\end{figure*}

NeuralVDB retains the VDB tree structure outlined above, but employs novel techniques to encode values, of both tiles and voxels, and topologies, of both nodes and the active states of values, cf.~active-masks mentioned in Section~\ref{sec:standard-vdb}. Whereas OpenVDB encodes values explicitly at full bit-precision, and NanoVDB (optionally) uses explicit but adaptive bit-precision, NeuralVDB instead uses neural representations for values, their states, and (optionally) parts of the tree-structure itself. Specifically, we are proposing two types of NeuralVDB that are optimized for respectively speed and memory. The first version, which we denote $[\textrm{Hash},5,4,\textrm{NN}(3)]$, only applies neural networks to the leaf nodes, whereas the second version is dubbed $[\textrm{Hash},5,\textrm{NN}(4),\textrm{NN}(3)]$ and applies neural networks heuristically to the two lower levels. As we shall demonstrate $[\textrm{Hash},5,4,\textrm{NN}(3)]$ favors fast random access whereas $[\textrm{Hash},5,\textrm{NN}(4),\textrm{NN}(3)]$ achieves a smaller memory footprint at the cost of slower access.

Our neural network architecture is based on several multi-layer perceptrons (MLPs) that partition the entire coordinate span of the sparse volume into partially overlapping domains (more on this partitioning in Section~\ref{sec:domaindecomp}).
Each MLP maps floating-point voxel coordinates $(x,y,z)$ to the relevant value type of the VDB \ken{tree}, \doyub{e.g.,~scalar, vector, and \ken{binary} mask values. For the scalar and vector values, we use the MLP as \ken{a} regression network. \minjae{We encode the} \ken{binary} mask, which indicates whether a given coordinate \ken{maps to an} active value/child or not, using an MLP classifier. We will \ken{cover the} details \ken{of} this classifier network in Section~\ref{sec:encoding-hierarchy}. }

The regression MLPs are defined through training, which optimizes 
a mean squared error (MSE) loss function of the type

\begin{equation}
\label{eq:mse}
L_{MSE}(f, \hat{f}) = \frac{1}{N} \sum_{i=1}^{N}(f - {\hat{f}}_i)^2
\end{equation}
where $f$ is the target value and $\hat{f}$ is the predicted value from the network. For an SDF data, we scale the target to be in the range of $[-1, 1]$, whereas for the fog volumes, we keep the original range, which is \kenmr{typically} $[0, 1]$. \doyub{For the classification MLPs, we use cross-entropy loss.} We \kenmr{also use} stochastic gradient descent with an Adam optimizer~\cite{kingma2014adam}. Learning rate is scheduled to decay exponentially for every epoch. In Section~\ref{sec:results}, we list all the hyperparameters that we used to perform the experiments.

While training of MLPs is occasionally straightforward, it is well-known that in many practical applications MLPs often fail to reconstruct high-frequency signals, even with high-capacity, i.e.,~wide/deep, networks~\cite{jacot2018neural}. We apply two different techniques to mitigate this issue: Firstly we restrict the training samples to active values only, and secondly we map the low dimensional feature $(x, y, z)$ to different feature spaces for better accuracy. We will elaborate more on both these ideas below.

\subsubsection{Sparse Field Training}
\label{sec:sparse-field-training}

\begin{figure*}[t]
    \centering
    \includegraphics[width=\textwidth]{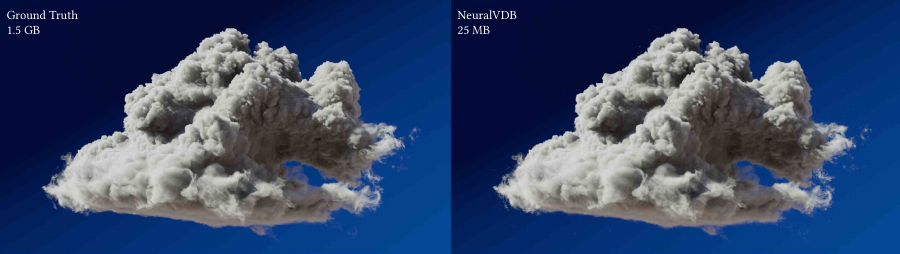}
    \caption{Examples of reconstructed volumes from NeuralVDB on Disney Cloud dataset~\cite{disneyclouds}}
    \label{fig:hero}
\end{figure*}

The encoding process of the value regression MLP starts with an existing VDB grid, either represented as an OpenVDB or NanoVDB. For each epoch, i.e.,~pass over the training set, we randomly sample the \textbf{active} voxels, thus explicitly excluding all inactive values, e.g.,~background values, encoded in the VDB tree \minjae{since,} by design, active values are used to indicate that a value is significant. This is a simple but efficient way to introduce sparseness in the training despite the fact that tree nodes are dense. For instance, a narrow-band level set is represented as a truncated signed distance field where the active voxels ``uniformly sandwich'' the zero-crossing surface, i.e.,~a narrow-band level set of width six has active voxels in the range [$-3\Delta, 3\Delta x$] where $\Delta x$ denotes the size of a voxel. Conversely, a fog, i.e.,~normalize density, volume typically has a wider active value set, but they are still sparse in the sense that the active set is bounded, typically with non-trivial boundaries, e.g.,~see the cloud example from Figure~\ref{fig:hero}. Training a network with only these active voxels allows the model to focus its learning capacity on the most important content encoded into a VDB tree; thus the adaptive structure of VDB is encoded implicitly into the network during training. \doyub{The effect of training with sparsity information is demonstrated in Appendix~\ref{appx:effect-of-training-with-sparsity}}. Obviously, this network alone will not extrapolate well outside the active voxels, which is by-design. Therefore, the hierarchical structure from the source VDB is embedded as part of the NeuralVDB data, except the dense leaf nodes, to mask out any random access outside the active voxel regions which are not trained.

\subsubsection{Feature Mapping} 
\label{sec:feature-mapping}
As shown in recent work on spectral bias and Neural Tangent Kernels~\cite{rahaman2019spectral,jacot2018neural,tancik2020fourfeat}, a vanilla MLP tends to fail to capture high-frequency details even with deep and wide networks. It was demonstrated in~\cite{jacot2018neural} that the effective regression kernel width of a regular MLP is too wide to represent such signals. To overcome this issue, a number of different techniques have been proposed, including positional encoding~\cite{mildenhall2020nerf}, and Fourier feature mapping (FFM)~\cite{tancik2020fourfeat} as its generalization. Different mapping techniques have been proposed from different contexts as well such as one-blob encoding~\cite{muller2019neural,muller2020neural}, triangle wave~\cite{mueller2021realtime}, or neural hash encoding~\cite{mueller2022instant}. These mapping (or encoding) techniques transform input coordinates, $\mathbf{x} \in \mathbb{R}^3$, into higher dimension vectors $\gamma(\mathbf{x})$

\begin{equation}
\label{eq:fm}
\mathbf{z} = \gamma(\mathbf{x})
\end{equation}
where $\mathbf{z} \in \mathbb{R}^{n}$ and $n \gg 3$ where $n$ is the new feature dimension. By applying such mappings, an MLP can converge faster with fewer parameters and shorter training times. Alternatively, the domain itself can be decomposed into smaller geometrical representations, such as octrees~\cite{takikawa2021neural} or grid of subdomains~\cite{moseley2021finite}, which tackles the spectral bias problem, i.e., the fact that networks tend to bias towards low frequency signals in the training set. However, we prefer feature mapping techniques over the geometric approaches to decouple the neural network design from the VDB tree structure. This way, the architecture is open to other feature mapping methods such as neural hash grids~\cite{mueller2022instant} and can adopt new techniques without heavy refactoring. Therefore, we implement FFM as the main feature mapping method in the NeuralVDB framework. 



The final NeuralVDB data is then a concatenation of mask-only VDB trees with the value regressor MLP network (see Figure~\ref{fig:overview}). While this already reduced the memory footprint significantly (see Table~\ref{tab:memory_cost}), we show that the memory efficiency can be further improved by encoding the hierarchy of the VDB tree with neural networks in the following section.

\subsection{Hierarchical Networks}
\label{sec:hierarchical-net}

\begin{table*}[t]
    \centering
    \resizebox{1.0\textwidth}{!}{
        \input{tables/memory_cost}
    }
    \caption{Memory Cost of the VDB Hierarchy. Columns on the left (Standard VDB) show the statistics of the dragon model represented in standard VDB format. Columns in the middle (NeuralVDB $[\textrm{Hash},5,4,\textrm{NN}(3)]$) show similar statistics when only the voxel values are encoded in neural networks. The right-most columns (NeuralVDB $[\textrm{Hash},5,\textrm{NN}(4),\textrm{NN}(3)]$) shows the numbers when neural networks are used to encode both tree hierarchy and values for two lower levels. NeuralVDB with $[\textrm{Hash},5,4,\textrm{NN}(3)]$ was able to reduce its size down to 6.268\% of the original VDB and NeuralVDB with $[\textrm{Hash},5,\textrm{NN}(4),\textrm{NN}(3)]$ achieved even smaller footprint. Note that due to the sparse domain decomposition described in Section~\ref{sec:domaindecomp}, the voxel values are encoded with multiple neural networks where each network encodes its dedicated bounding box.}
    \label{tab:memory_cost}
\end{table*}

\begin{figure*}[t]
\includegraphics[width=\textwidth]{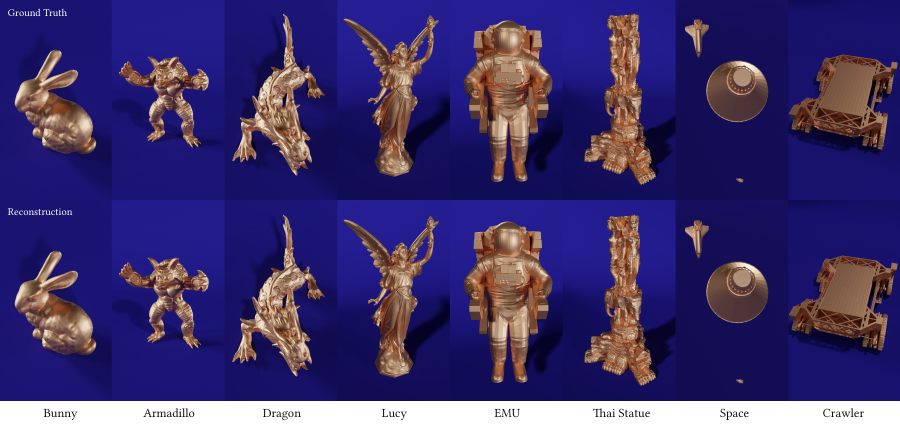}
\caption{Ground truth SDF VDB models (top row) and reconstructed SDF VDB models using NeuralVDB (bottom row).}
\label{fig:sdf-results}
\end{figure*}

As indicated above, NeuralVDB achieves a significant reduction in its memory footprint, relative to OpenVDB, by replacing dense tree nodes with a shared neural network. To motivate some of our design decisions consider Table~\ref{tab:memory_cost}, where we quantify this memory reduction for a specific sparse volume, namely the level set model of the dragon shown in third column of Figure~\ref{fig:sdf-results}. This table shows node counts and memory footprints at different tree levels for one standard and two neural representations with the same low reconstruction error (Intersection over Union (IoU) of $99\%$). The first column, with OpenVDB, denoted $[\textrm{Hash},5,4,3]$, clearly shows that the overall memory footprint is dominated by the voxels, i.e.,~values in the leaf nodes, that take up $94\%$ of the total footprint. The neural representation of voxels, shown in the middle column and denoted $[\textrm{Hash},5,4,\textrm{NN}(3)]$, reduces the footprint of the leaf values to only $6\%$, corresponding to $16\times$. However, the total footprint is now dominated by the leaf bit masks and the internal nodes at level 1, i.e.,~the states of the voxels and the nodes just above the leaf nodes. \doyub{As stated in Section~\ref{sec:introduction}, one of our key design goals is to preserve} \ken{as much of the information captured in the source VDB data structure as possible, which includes the hierarchical tree structure as well as the spatial occupancy,~i.e., topology, and values of the sparse volumetric data. In other words, we seek a more compact neural representation of the source tree structure that encodes most if not all of its payload. A natural approach} is therefore to apply neural representations to all voxels, as well as their masks and parent nodes, which is shown in the right-most column of Table~\ref{tab:memory_cost}, denoted $[\textrm{Hash},5,\textrm{NN}(4),\textrm{NN}(3)]$. This results in an overall compression factor of $68\times$ when comparing $[\textrm{Hash},5,4,3]$ at $257MB$ to $[\textrm{Hash},5,\textrm{NN}(4),\textrm{NN}(3)]$ at $3.8MB$. Note that we use the same network capacity for the voxels and masks at level 0, resulting in virtually identical footprints. Interestingly, the neural compression of the two lowest levels of the VDB tree structure results in a hierarchical representation, $[\textrm{Hash},5,\textrm{NN}(4),\textrm{NN}(3)]$, whose memory footprint is still dominated by those two lowest levels. This seems to suggest that neural representations of the remaining top levels, $2$ and $3$, will have little impact on the overall memory footprint.

\subsubsection{Encoding Hierarchy}
\label{sec:encoding-hierarchy}

Based on the observations above, we propose only to introduce hierarchical neural networks at the two lowest levels of VDB tree structure. More precisely, we replace voxel and tile values at levels 0 and 1 with MLP-based \textbf{value regression} networks as well as child and active masks at level 1 and active masks at level 0 with \textbf{classifiers}. The root and upper internal levels of the tree structure shall remain unchanged. This configuration is illustrated in the right column of Figure~\ref{fig:overview}. The mask classifier at level 1 is trained with level 1 child nodes' coordinates as the input and its child and active masks $m_1 \in \left \{ c_1 = 1, a_1 = 1 \right \}$ as the target labels. Thus, this ternary classifier predicts three possible cases, 1) a leaf child node, 2) an active tile value, or 3) an inactive tile value, from the input coordinates. Conversely, the classifier at level 0 is trained with voxel coordinates as the input and the active leaf masks $m_0 \in \left \{ a_1 = 1 \right \}$ as the label. Thus, this binary classifier predicts whether given coordinates map to active or inactive voxels. To optimize the parameters, cross-entropy loss is used for the level-1 mask classifier and binary cross-entropy (BCE) loss is used for the level-0 mask classifier. For the nodes at level 1 with tile values ($m_1 = 0$), these tile values are also encoded using an MLP-based value regressor, similar to the voxel value regressor.

Note that the level-0 mask classifier is essentially an occupancy network. When reconstructing voxel occupancy, the BCE loss function can be tweaked to tackle sparse and imbalanced distribution as well as the vanishing gradient problem~\cite{brock2016generative,saito20183d}. However, the level-0 mask network is performed within level-1's chidren nodes which addresses the imbalance problem since the children nodes are allocated only around where the actual values are, instead of its full domain. Also, a typical network depth is not very deep (e.g., [2, 4]), and hence gradients do not vanish easily. Therefore, we keep the vanilla BCE without further tuning.

Due to the  hierarchical nature of the tree structure, the capacities of mask classifier and tile value regressor at level 1 are typically much smaller than the capacities of the mask classifier or voxel regressor at level 0. During the reconstruction, we perform top-down traversal by first querying the level-1 mask classifier. If the query point is classified as an active tile, then the corresponding tile value is predicted and returned using the tile value regressor. Conversely, if the query is classified as a leaf node, its mask classifier is used to determine the active state. The query points that map to active states are then used for the final inference through the value regressor, mimicking the tree traversal/early termination of the standard VDB tree.

\subsubsection{Source Embedding} 
\label{sec:source-embedding}

Although the networks with FFM~\cite{tancik2020fourfeat}, which is our feature mapper of choice as mentioned in Section~\ref{sec:feature-mapping}, can classify level-1 and voxel masks accurately; it still might produce a number of positive samples that are incorrectly classified. However, we observed that the number of such samples is relatively small (e.g., <1\% of all positive samples for level-1 masks and ~5\% of active voxel masks), and in fact, can be appended to the data structure.

For the active mask classifier for voxels, however, even a percent of false positives might result in significant number of voxels to embed since the number of active voxels easily exceeds tens of millions (see Table~\ref{tab:non-temporal-results} and \ref{tab:temporal-results}). While this is impractical and defeats the purpose of space efficiency, most of such false negatives are near the decision boundaries (not the geometrical boundaries). Based on this observation, we filter out voxels that are far enough from the surface (in case of SDF) or do not have significant value (in case of volume density or any other scalar fields).  This remedy seems to work well enough not to show any significant artifacts.

\begin{figure*}[t]
    \centering
    \includegraphics[width=\textwidth]{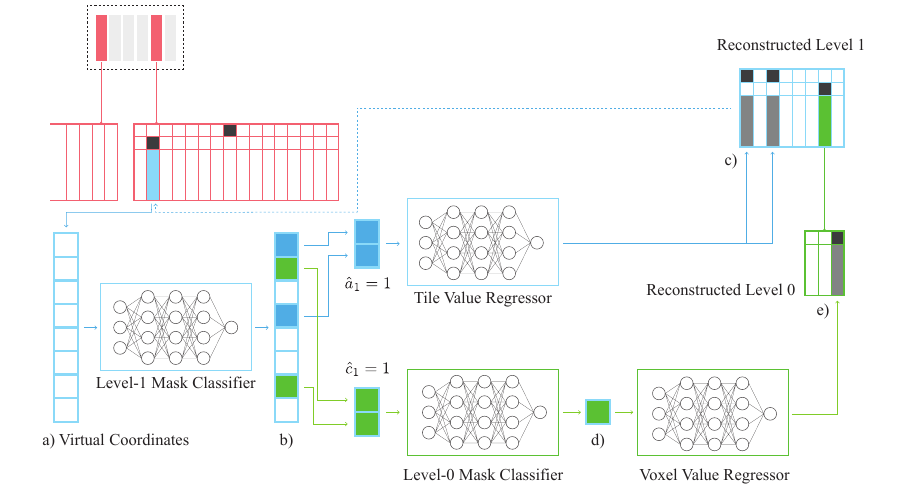}
    \caption{Reconstructing VDB from a NeuralVDB data. Virtual coordinates from level 1 are classified into either one of 1) child node, 2) active tile, 3) inactive tile. From the resulting vector at b), active mask coordinates are then further passed down to the tile value regressor to reconstruct the tile values at c). Input coordinates with child mask on are passed to the level-0 mask classifier to check active voxel state and then active voxel d) is finally used to infer the voxel value for the reconstruction of level 0 at e). }
    \label{fig:reconstruction}
\end{figure*}

\subsection{Sparse Domain Decomposition}
\label{sec:domaindecomp}

When a scene is too large and/or contains disjoint clusters of volumes, a single network can perform poorly since the input coordinates are normalized between $[0, 1]$ before the feature mapping stage. In contrast, the value-mapping in a standard VDB is agnostic to such an incoherent clustering of voxels. To address the problems above, we propose a sparse domain decomposition approach, which is inspired by the sparsely-gated Mixture-of-Experts (MoE) method~\cite{shazeer2017outrageously}. First, we decompose the domain with fixed-size subdomains where each subdomain $D_k$ spans configurable size in index space in range of 512 to 2048. A subdomain has a fixed-width halo that overlaps with other adjacent subdomains. We chose 8 voxels for the halo size which is wide enough to eliminate the discontinuity and small enough to reduce the compute overhead. \ken{The entire domain is partitioned into a regular grid of subdomains, where empty subdomains are discarded.} Also, a dedicated neural network (expert) is defined for each subdomain. For simplicity, the same network architecture is used for all the experts. Given this setup, we define a gate function $G(\mathbf{x})_k$ for each subdomain $D_k$\ken{, where} \doyub{$\mathbf{x}$ is a normalized coordinates between $[0, 1]$ for the given subdomain bounding box}. This gate function $G(\mathbf{x})_k$ is defined as a clamped tent function (a tent function with max value of 1 uniformly outside the overlapping region) which covers the subdomain $D_k$ including the halo. When input coordinates are passed, the gate functions and the expert networks generates the output as

\begin{equation}
\label{eq:moe}
\hat{y} = \sum_{k=1}^n G(\mathbf{x})_k E_k(\mathbf{x})
\end{equation}
where $n$ is the number of subdomains and output $\hat{y}$ can be one of the child/active masks or voxel values, which means the sparse subdomain decomposition can be applied to any neural modules in our framework (see Figure~\ref{fig:overview}b for the reference). Note that the gate function above is not learnable, which is different from the sparsely-gated MoE~\cite{shazeer2017outrageously}. Also, a single input coordinate can activate (return non-zero output) multiple gate functions (as many as eight) due to the overlapping halos, and we average the evaluated values weighted by the gate functions. \doyub{In practice, we examine the gate function first to determine which network should be invoked and only perform the computation for the networks with non-zero gate values.}
Since each subdomain has dedicated classifiers and regressors, we can train concurrently on multiple GPUs. When multiple GPUs are used, groups of subdomains (since there can be more subdomains than number of available GPUs) are assigned for each GPU. After training, the groups of the subdomains are merged into a single NeuralVDB structure. 

Using the sparse domain decomposition outlined above, large sample scenes like the Space model with a voxel resolution of $32844\times 24702\times 9156$ in Figure~\ref{fig:sdf-results}, can be effectively handled without sacrificing accuracy. In this particular case, twelve subdomains for the entire scene are allocated in total by our algorithm \doyub{(i.e., subdividing the entire domain into a grid of subdomains and discarding the subdomains without any voxels)}. The sizes are determined heuristically as described in Appendix~\ref{appx:hyperparameters}. 

\subsection{Reconstruction}

\begin{figure}
\centering
\includegraphics[width=\linewidth]{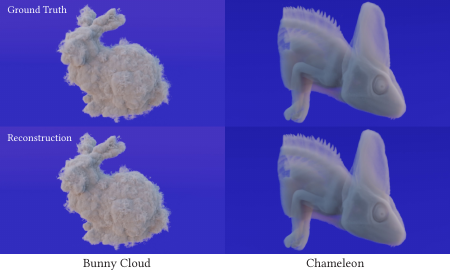}
\caption{Ground truth volume density VDB models (top row) and reconstructed volume density VDB models using NeuralVDB (bottom row). The Chameleon model is acquired from Open Scientific Visualization Datasets where the original dataset is from DigiMorph~\cite{chameleondataset}.}
\label{fig:vol-results}
\end{figure}

So far we have focused on how standard VDB trees can be compactly \textbf{encoded} in NeuralVDBs by means of training various neural networks. This of course leaves the problem of efficiently \textbf{decoding} NeuralVDBs by inferencing, which is the topic of this section. We will consider two fundamentally different scenarios. First, we show how a standard VDB can be reconstructed from an existing NeuralVDB representation, which is useful when a NeuralVDB is stored offline, e.g.,, on disk or transmitted over a network, and needs to be decoded into a standard VDB in memory. This is typically an offline process where we reconstruct the entire VDB tree in a single sequential pass thought the NeuralVDB data. Second, we show how we can support random access to values directly from in-memory NeuralVDB data, without first fully reconstructing the entire VDB tree. The first case favors memory efficiency over reconstruction time, whereas the latter needs to balance these two factors in order to allow for reasonable access times for applications like rendering and collision detection. To this end we propose the two different configurations of NeuralVDB, $[\textrm{Hash},5,4,\textrm{NN}(3)]$ and $[\textrm{Hash},5,\textrm{NN}(4),\textrm{NN}(3)]$ introduced Section~\ref{sec:hierarchical-net}. We will elaborate more on these two cases below.

\paragraph{Offline Sequential Access}

For applications that prioritize a low memory footprint over fast reconstruction times, we use the NeuralVDB configuration denoted $[\textrm{Hash},5,\textrm{NN}(4),\textrm{NN}(3)]$. Examples of such applications are storage on slow secondary-storage devices like hard drives and DVDs or transfer over low-bandwidth internet. The reconstruction into a standard VDB tree only requires a single sequential pass over the compressed data. Since the root and its child nodes are encoded identically to a standard VDB tree, we will limit our description of the reconstruction to the lower two levels of the tree that use neural representations. Sequential access to level 1 nodes is straightforward since their coordinates are trivially derived from the child masks at level 3 (see \cite{museth2013vdb} for details on how bit-masks compactly encode coordinates). Thus, for each node at level 1 (of size $16^3$) we use standard inference to reconstruct the child and active masks from the classifiers and the tile values from the value regressors described in Section~\ref{sec:encoding-hierarchy}. We correct the masks with the list of false positives that we explicitly encoded during the training step (see \ref{sec:vdb-with-nn}). Next, using the child masks at level 1 we proceed to visit all the leaf nodes (of size $8^3$) and sequentially infer the voxel values and their active states from the value regressor and binary classifier at level 0. \doyub{During the decoding process, we use disjoint blocked ranges, \ken{which are distributed amongst multiple GPUs and subsequently merged} into a single output VDB. Since each blocked range has dedicated classifiers and regressors, like in \ken{the} training stage, inferencing can \ken{also} be performed concurrently on multiple GPUs. When reconstructing one of these blocked ranges, it still has access to all the networks, meaning it can still reconstruct volumes without discontinuity thanks to Equation~\ref{eq:moe}}.

\begin{figure*}[t]
    \includegraphics[width=\textwidth]{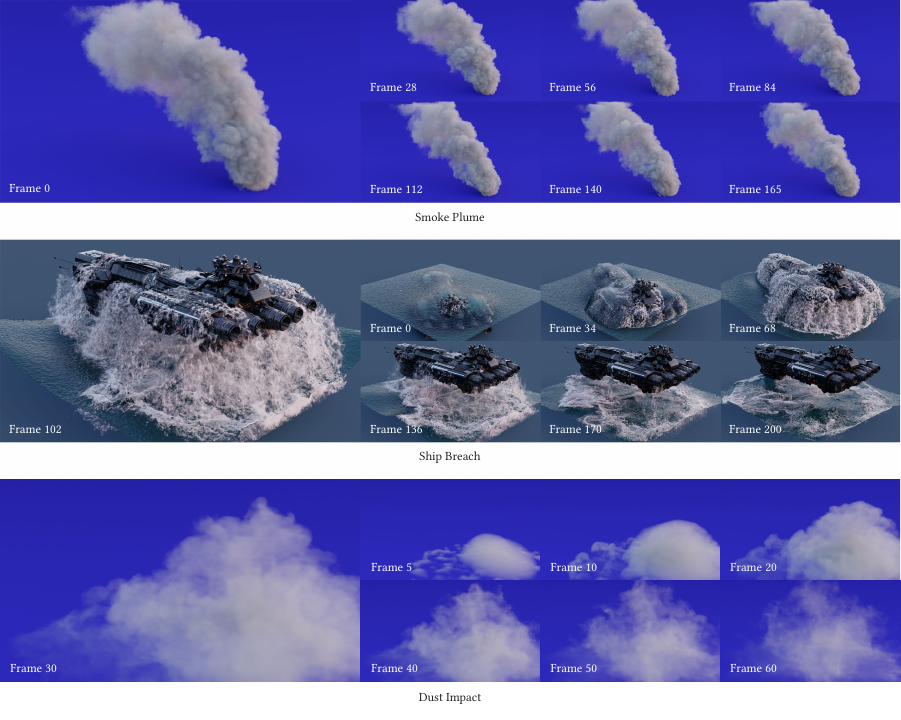}
    \caption{Reconstructions from temporally encoded NeuralVDB examples. Smoke Plume is simulated density volumes for $0-165$ frames, Ship Breach is signed distance fields of a spaceship breaching a water surface for $0-200$ frames, and Dust Impact is simulated density volumes for $0-166$ frames. }
    \label{fig:temporal-results}
\end{figure*}

\begin{figure}
    \centering
    \includegraphics[width=\linewidth]{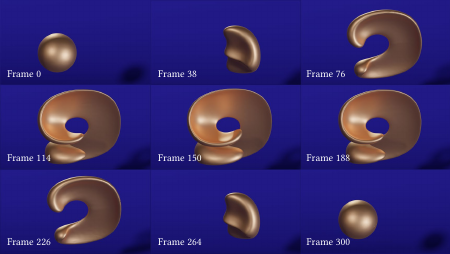}
    \caption{Reconstruction from temporally encoded NeuralVDB data for $0-300$ frames, procedurally generated based on LeVeque's Test~\cite{leveque1996test} (also known as Enright test).}
    \label{fig:leveque-results}
\end{figure}

\begin{figure}
    \centering
    \includegraphics[width=\linewidth]{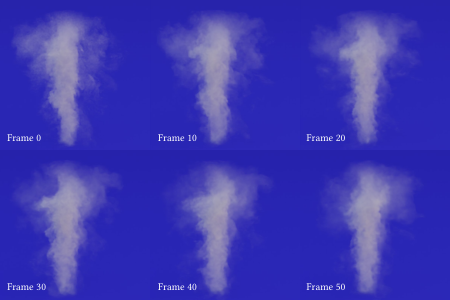}
    \caption{Reconstruction from temporally encoded NeuralVDB data from simulated density volumes for $0-127$ frames, dataset generated from EmberGen Tornado simulation~\cite{embergendataset}.}
    \label{fig:tornado-results}
\end{figure}

\paragraph{Online Random Access}

Since $[\textrm{Hash},5,\textrm{NN}(4),\textrm{NN}(3)]$ employs hierarchical neural networks (two levels) we have found this configuration to be too slow for real-time random access applications. Consequently we propose $[\textrm{Hash},5,4,\textrm{NN}(3)]$, show in the left column of Figure~\ref{fig:overview}, for applications that require both fast random access and a small memory footprint since it uses the proven acceleration techniques of VDB for the tree traversal in combination with the compact neural representation of the voxel values only. In other words, random access into $[\textrm{Hash},5,4,\textrm{NN}(3)]$ has the same performance characteristics as a standard VDB tree, except for leaf values that require an additional regression for the voxels. As shown in the middle column in Table~\ref{tab:memory_cost}, $[\textrm{Hash},5,4,\textrm{NN}(3)]$ still has an in-memory footprint that an order of magnitude smaller than $[\textrm{Hash},5,4,3]$. While $[\textrm{Hash},5,4,\textrm{NN}(3)]$ consumes more memory than $[\textrm{Hash},5,\textrm{NN}(4),\textrm{NN}(3)]$, it still benefits from a massive compression ratio of the leaf level value regression network. Moreover, $[\textrm{Hash},5,4,\textrm{NN}(3)]$ can be trivially reconstructed from the other version, $[\textrm{Hash},5,\textrm{NN}(4),\textrm{NN}(3)]$, by leaving the voxel regressor unchanged, and can therefore be seen as a pre-cached representation for the faster access, similar in spirit to \cite{hedman2021baking}. Once the $[\textrm{Hash},5,4,\textrm{NN}(3)]$ representation is available, random access becomes a simple two-step process: 1) Use standard (accelerated) random access techniques (see~\cite{museth2013vdb}) to decide if a query point maps to a tile or a voxel, i.e.,~level $\{3,2,1\}$ or ${0}$. 2) if it is a tile, return the value explicitly encoded into the standard VDB structure, and else, predict the voxel values using the regressor.

While $[\textrm{Hash},5,4,\textrm{NN}(3)]$, the in-memory representation of NeuralVDB, can be viewed as a cached evaluation of offline representation $[\textrm{Hash},5,\textrm{NN}(4),\textrm{NN}(3)]$, there are still room for more active caching mechanism such as caching of evaluated voxel masks/values in a cyclic buffer to reduce number of neural network inferences. We are investigating this approach as part of our future work.

\subsection{Temporally-Coherent Warm-Start Encoder}

One of the main sources of sparse volumetric data are simulations. As such, one of the key applications for OpenVDB, and hence by extension NeuralVDB, is time-sequences of animated sparse volumes. This presents both an opportunity for acceleration as well as a challenge in terms of expected temporal coherence. We achieve both of these with a relatively simple idea, namely that of warm starting the neural training, i.e.,~encoding, of one frame with the converged network weights from the previous frame. As indicated, this has two significant benefits that are unique to NeuralVDB. Firstly, the coupling (through initialization) to a previous frame introduces temporal coherency across frames, and secondly it accelerates the training times, typically by a factor of $1.5-2.5$ times, when compared to a ``cold-start'' training. Thus, our novel warm-start encoder leverages temporal coherency of the input volumes to preserve temporal coherency of the output volumes (see Figure~\ref{fig:temporal-results}), in addition to reducing encoding times (see Section~\ref{sec:results}). Specifically, we run the encoder sequentially from the first frame to the last frame, while saving neural networks per frame to re-use them in the following frame as a warm-starter to achieve temporally coherent network weights. If the input volumes contain high-frequency details, like thin layers of smoke, then a naive (``cold-start'') encoding can produce flickering due to the fact that a fixed learning rate for all frames can introduce discontinuities of network weights across frames. In order to fix the issue, we run the first frame with the target learning rate, and re-process the first frame with the same or smaller learning rate (e.g., up to 100 times smaller). The rest of the frames are processed only once using the new learning rate, and this step reduces the training iteration when the loss becomes lower than the first frame's final loss. This technique is similar to the fine-tuning method for transfer learning~\cite{zhou2017fine} where it adapts to the new target (new frame) without drifting too much from the old target (previous frame). When the domain decomposition step adds a new domain in the middle of the animated sequence, we repeat the same process of encoding the domain with the target learning rate, then again with the smaller one. Warm starting not only produces temporally coherent results but also boosts encoding performance while satisfying both quality and compression ratio requirements as shown in Table~\ref{tab:temporal-results}.

%% file: tables/memory_cost.tex
\begin{tabular}{rrrrrrrrrr}
\toprule
                   & \multicolumn{2}{l}{Standard VDB} & \multicolumn{3}{l}{NeuralVDB ([\textrm{Hash},5,4,\textrm{NN}(3)])} & \multicolumn{4}{l}{NeuralVDB ([\textrm{Hash},5,\textrm{NN}(4),\textrm{NN}(3)])} \\
                   &   Num. Nodes &       Bytes &                                     Num. Nodes &  Params &      Bytes &                                                  Num. Nodes &  Params & Patches &     Bytes \\
\midrule
Internal (Level 2) &            8 &     327,776 &                                              8 &         &    327,776 &                                                  8 &         &         &   327,776 \\
Internal (Level 1) &          318 &   5,539,560 &                                            318 &         &  5,539,560 &                                                318 &  99,332 &   1,879 &   423,692 \\
    Mask (Level 0) &      124,166 &   9,436,616 &                                        124,166 &         &  9,436,616 &                                                    & 395,268 &   7,293 & 1,668,588 \\
  Voxels (Level 0) &   63,572,992 & 254,291,968 &                                                & 398,352 &  1,593,408 &                                                    & 395,268 &         & 1,581,072 \\
             Total &              & 269,595,920 &                                                &         & 16,897,360 &                                                    &         &         & 4,001,128 \\
                   &              &             &                                                &         &     6.268\% &                                                    &         &         &    1.484\% \\
\bottomrule
\end{tabular}

%% file: 4_results.tex
\input{tables/numbers}

In this section, we test NeuralVDB under a number of scenarios, including encoding, decoding, and random access. All the numerical experiments were performed on a virtual machine with NVIDIA RTX A40 GPUs and a host AMD EPYC 7502 CPU. NeuralVDB is implemented in C++17 and makes use of both CUDA and PyTorch~\cite{NEURIPS2019_9015}.

\begin{figure}
\centering
\includegraphics[width=\linewidth]{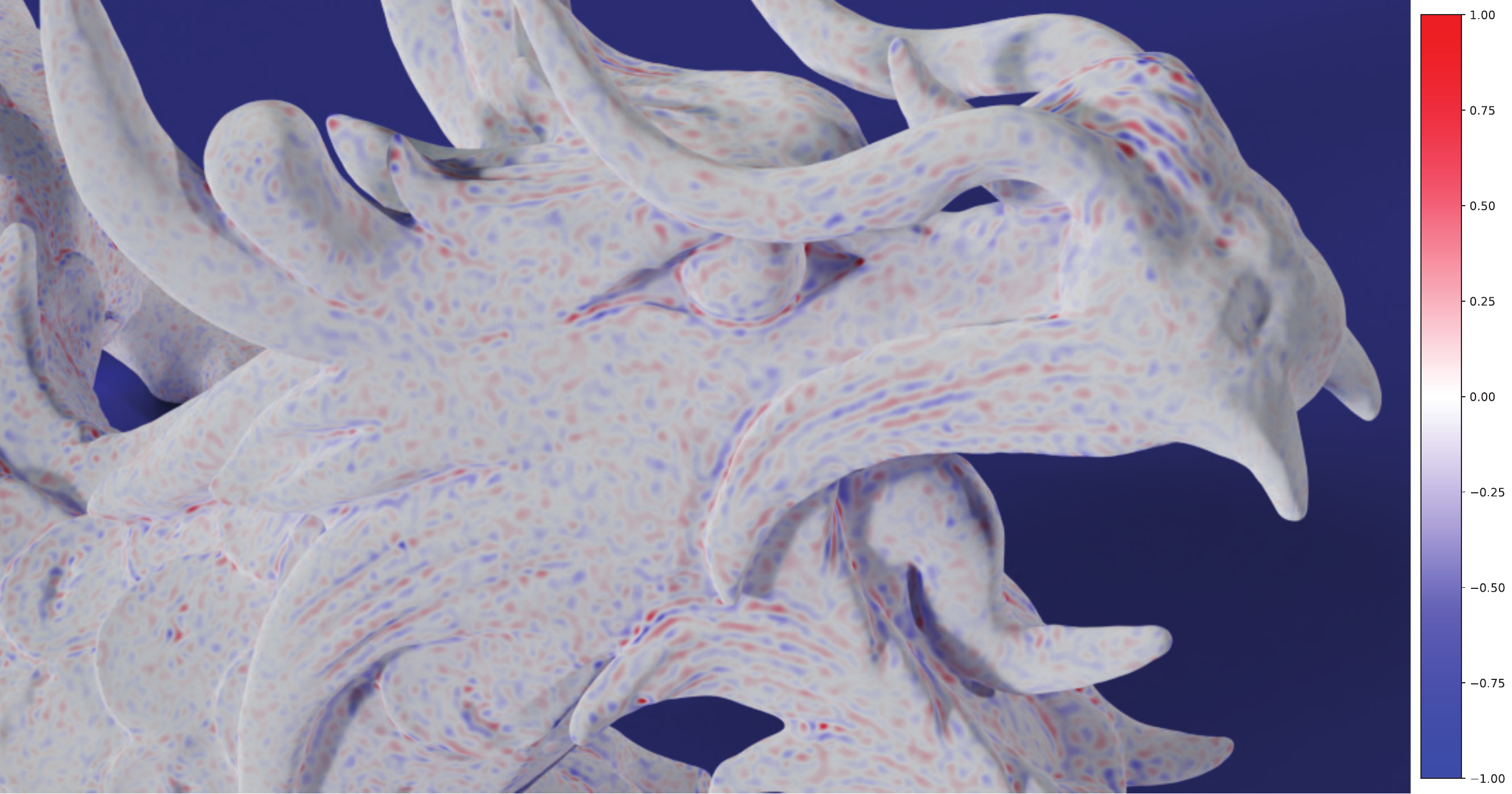}
\caption{The offset between the ground truth and reconstructed meshes are rendered with a color map. Red and blue indicate positive and negative displacements relative to the outward normal direction. Unit of the color map is the voxel size of the source VDB grid.}
\label{fig:sdf-error}
\end{figure}

\begin{figure}
\centering
\includegraphics[width=\linewidth]{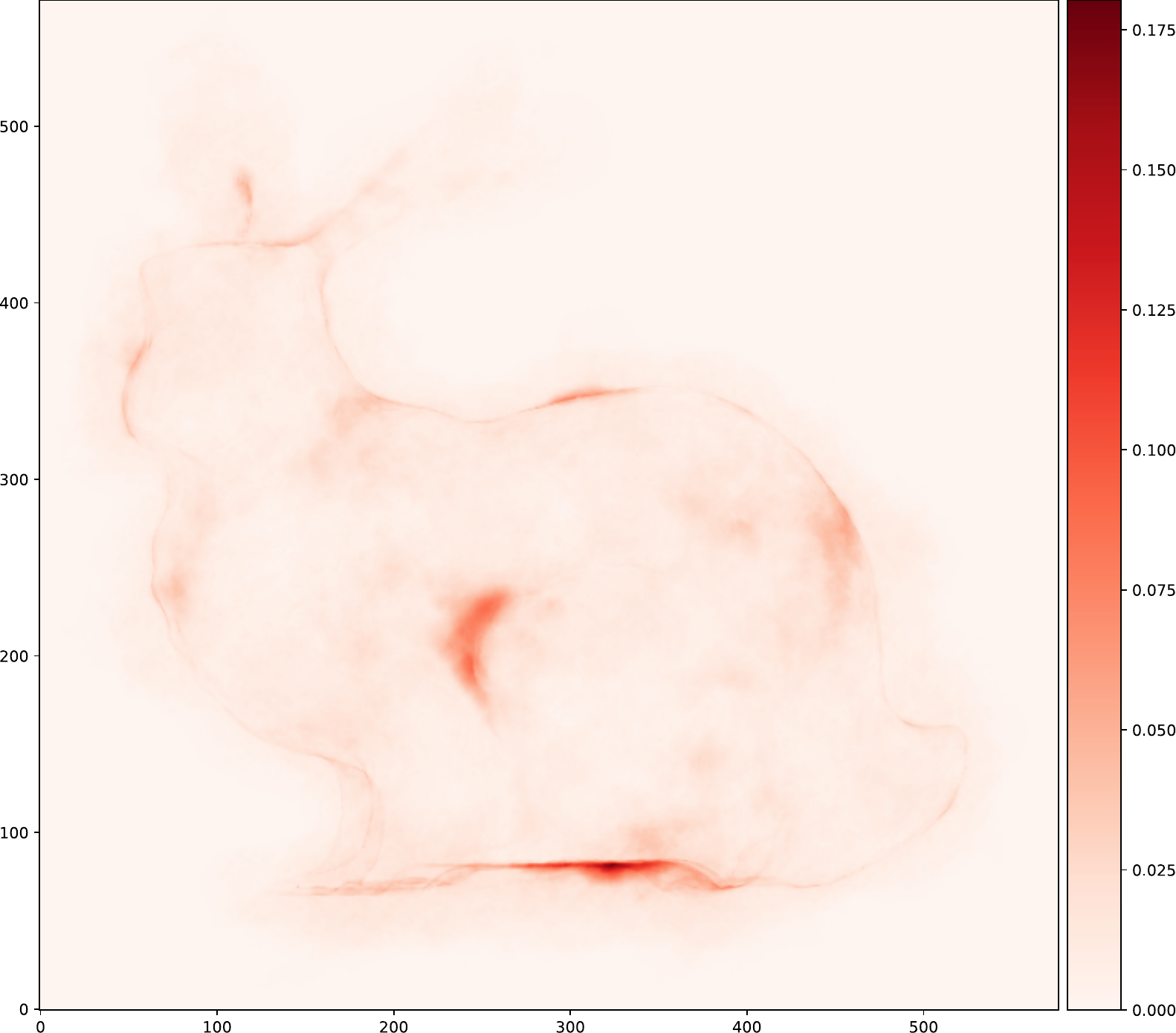}
\caption{Error visualization for the Bunny Cloud example. The absolute error is averaged in z-axis.}
\label{fig:vol-error}
\end{figure}

\begin{figure*}[t]
\centering
\includegraphics[width=\textwidth]{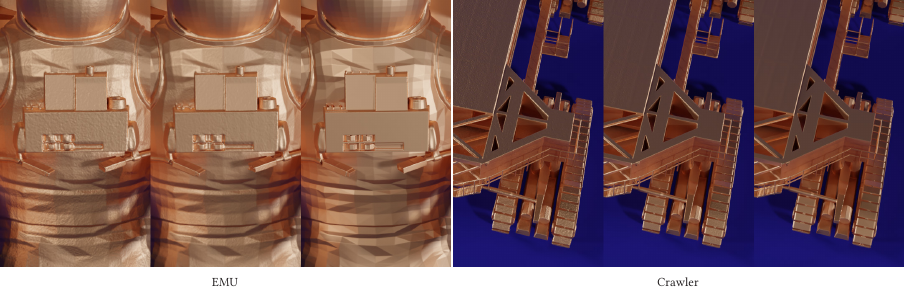}
\caption{Visualization of the error convergence as more network parameters are used. For each example, the left-most column corresponds to the baseline reconstruction where fewer parameters are used. The center column shows the result from a larger network (2$\times$ the width). The right-most column shows the ground truth. For the EMU example, the compression ratio is \EMUCOMPRATIO and \EMUHIGHRESCOMPRATIO for the smaller and larger models, respectively. For the Crawler example, the compression ratio is \CRAWLERCOMPRATIO and \CRAWLERHIGHRESCOMPRATIO for the smaller and larger models.}
\label{fig:errors}
\end{figure*}

\subsection{Encoding}
We first \kenmr{evaluate our new VDB architecture by analyzing its efficiency at encoding a variety of model volumes with a given} quality criteria expressed as specific error tolerances. We define our main target error metric to be Intersection of Union (IoU) for narrow-band level sets, i.e.,~truncated signed distance fields (SDF), and Root Mean Squared Error (RMSE) for density volumes. Modified Chamfer Distance (mCD), which is a modified version of standard Chamfer Distance~\cite{wu2021density}, is also measured for level sets, which is defined as:

\begin{equation}
\label{eq:mcd}
\textup{mCD} = \frac{1}{2N_1}\sum_{i=1}^{N_1}SDF_2(\mathbf{v}_{1,i}) + \frac{1}{2N_2}\sum_{i=1}^{N_2}SDF_1(\mathbf{v}_{2,i})
\end{equation}
where the sampling points $\mathbf{v}_1\in V_1$ and $\mathbf{v}_2\in V_2$ were generated by extracting the isosurfaces from both ground truth ($SDF_1, V_1$) and the reconstructed VDBs ($SDF_2, V_2$). Note that the closest points to each other's surface are measured by directly sampling the SDF from the VDB data, which is different from the original Chamfer distance definition. \doyub{We acknowledge that relying solely on the mCD as a metric is insufficient, particularly because \kenmr{it was originally designed to evaluate}
point clouds~\cite{bouaziz2016modern}. Nevertheless, the mCD can still offer an indication of geometrical deviation when an implicit surface (SDF) is \ken{rendered as an explicit surface}. Hence, we enhance its assessment by incorporating IoU, following a similar approach to \ken{NGLOD}~\cite{takikawa2021neural}}. The hyperparameters were tuned to exceed 99$\%$ IoU for SDFs and produce an RMSE of less than 0.1 for the densities. Tables~\ref{tab:non-temporal-results} and \ref{tab:temporal-results} list the compression ratios for respectively non-temporal and temporal encoders. For the SDF models, $[\textrm{Hash},5,\textrm{NN}(4),\textrm{NN}(3)]$ achieved a compression ratio up to \textbf{61.2}, whereas for the density volumes, the compression ratio is as high as \textbf{140.9}. Figures~\ref{fig:sdf-results} and \ref{fig:vol-results} compare the ground truth with the reconstruction results of $[\textrm{Hash},5,\textrm{NN}(4),\textrm{NN}(3)]$. The Chameleon model achieved the best compression ratio among our dataset (140.9) since the data was smoother and evenly distributed compared to the other volumes. \ken{Consequently, the decision boundary of the classifier does not have to fit against high-frequency details, and the value regressor can use less neurons to represent a rather smooth value distribution.}

Figure~\ref{fig:leveque-results} shows reconstruction results from procedurally \kenmr{advected} SDFs called LeVeque's Test~\cite{leveque1996test}.
Figure~\ref{fig:temporal-results} and \ref{fig:tornado-results} shows simulation examples, Smoke Plume, Dust Impact, and Tornado from EmberGen VDB Dataset~\cite{embergendataset} and Ship Breach from the output of a high-resolution particle-based fluid solver. Table~\ref{tab:temporal-results} shows min, max, and mean values per column to illustrate variance of the temporal data.

While most of the compression ratios for the SDF volumes are in the range from 20 to 60, the Crawler model is an outlier in the sense that it only has a compression ratio of \textbf{13.3}. This particular SDF model is uniquely challenging because it contains some exceptionally thin geometric features as well as large flat surfaces. This amounts to both high- and low-frequency details, which are challenging to capture with a band-limited neural network. Consequently, this Crawler model requires a wider network with a higher capacity than most of the other SDF models, which in turn accounts for its lower relative compression radio.

\subsection{Reconstruction Error}
\label{sec:recon_error}
Given the fact that the proposed NeuralVDB representations are conceptually lossy compressions of standard VDB values (but importantly not its topology), it is \kenmr{essential} to investigate and understand the nature of these reconstruction errors.

In Figure~\ref{fig:sdf-error}, we visualize the error of the SDF reconstruction on the iso-surface mesh of the dragon model, by color-coding the closest distance to the ground truth. Specifically, the offset between the ground truth and the reconstruction is measured for each vertex of the reconstructed mesh. The blue-white-red color map shows the ``blobby'' error pattern generated by the NeuralVDB compression. This ``blobby'' error pattern is even more evident on flat surfaces, as shown in the two middle images of Figure~\ref{fig:errors} based on the spacesuit and Crawler SDF models. The right-most images in Figure~\ref{fig:errors} clearly show that this error can be significantly reduced by employing wider networks, of course at the expense of reduced compression ratios.

Finally, in Figure~\ref{fig:vol-error}, we compare renderings of the reconstructed density volumes relative to their ground truth representation. Small reconstruction errors are (barely) visible along the silhouettes in regions with small-scale details.

\subsection{Hyperparameters}

We have listed the hyperparameters used throughout this paper in Table~\ref{tab:hyperparameters}. Currently the capacity of the networks (number and width of the multiple  MLP layers) is chosen heuristically based on the complexity of the input volumes (more hidden neurons for more complex volume). Different activation functions are used for each example, based on heuristics discussed in Appendix~\ref{appx:hyperparameters}. For all the examples shown in Figure~\ref{fig:sdf-results} and \ref{fig:vol-results}, we use FFM~\cite{tancik2020fourfeat}.

\subsection{Performance}

As described in Section~\ref{sec:domaindecomp}, the sparse domain decomposition allows the encoding and decoding processes to be accelerated by multiple GPUs. In Table~\ref{tab:non-temporal-performance}, we report these speedup factors as a function of the number of GPUs applied to large volumes. For training, the subdomain resolutions listed Table~\ref{tab:non-temporal-results} and \ref{tab:temporal-results} are used. For reconstruction, \minjae{blocked ranges} of size $512^3$ are used for the job distribution onto multiple GPUs. As expected the strong-scaling is sub-linear, which is a consequence of the fact that both training and reconstruction have several sequential steps. This includes file I/O, domain decomposition, and gathering. Another factor that results in sub-linear strong-scaling is poor load balancing caused by imbalanced subdomains due to fluctuating sparse voxel counts in the subdomains. Still, Table~\ref{tab:non-temporal-performance} shows a significant benefit of using multiple GPUs for NeuralVDB. For certain combinations of input volumes and GPU counts, the automatic load balancers for the encoder and decoder determined that there are simply not enough subdomains \minjae{and/nor blocked ranges} to decompose and/or that using more GPUs is not beneficial. For instance, the Bunny model is smaller than the configured subdomain size (see Table~\ref{tab:non-temporal-results}). Also, the number of decoding \minjae{blocked ranges} (where each \minjae{range} has size of $512^3$) from the model is not large enough to use multiple GPUs. This decoding criterion is determined heuristically by checking if the number of average active voxels $\times$ number of \minjae{blocked ranges} is greater than or equal to 200 $\times$ number of GPUs.

The temporal warm-start encoder of NeuralVDB boosts performance of LeVeque's Test $2.4$ times, Smoke Plume $2.5$ times, Ship Breach $1.6$ times, Dust Impact $1.2$ times, and Tornado $3.1$ times. This is a significant benefit of warm starting each encoder with the converged neural network weights from the previous frame.

\begin{figure}
\centering
\includegraphics[width=\linewidth]{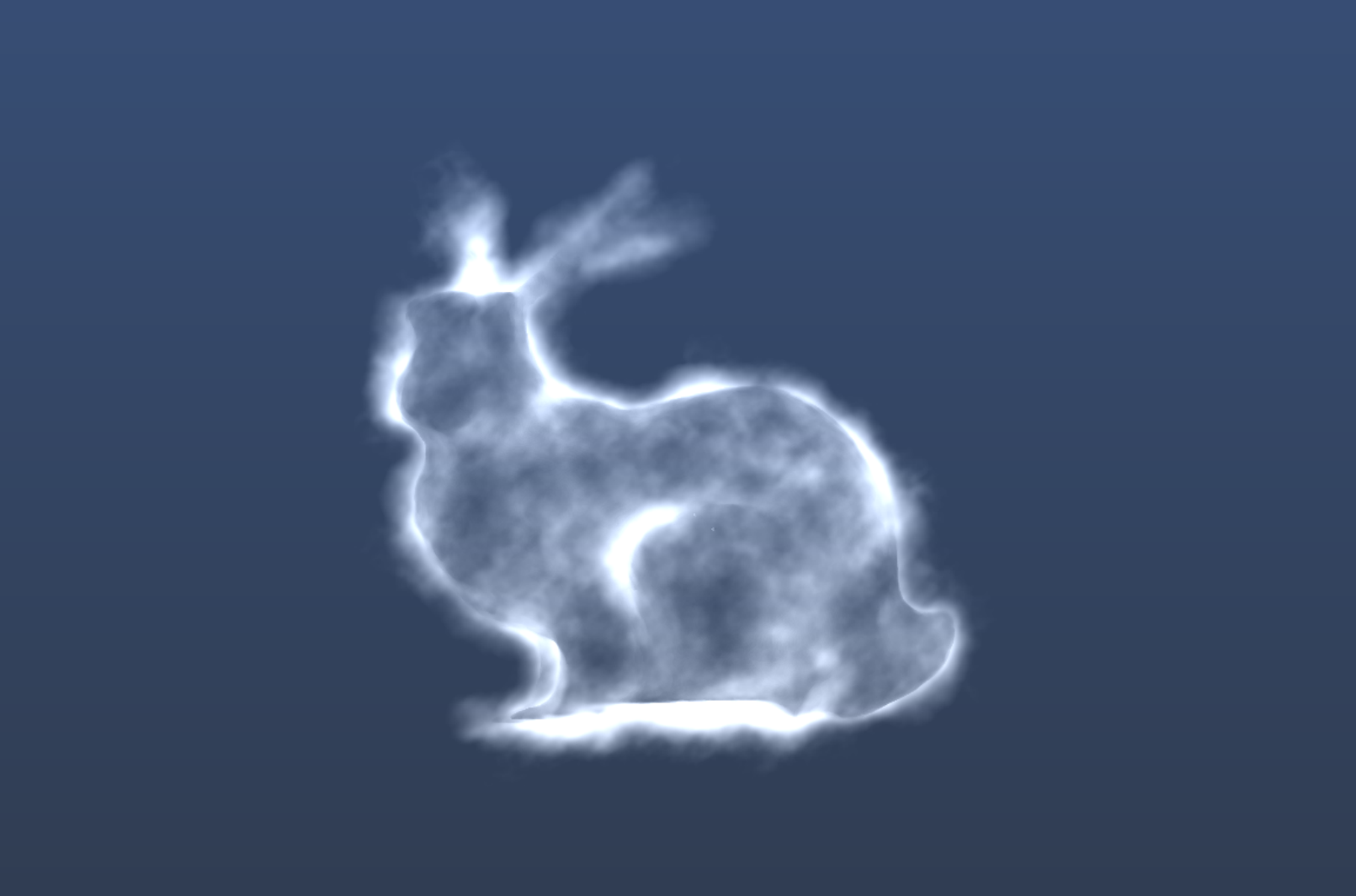}
\caption{Bunny Cloud model rendered with ray-marching directly on in-memory NeuralVDB.}
\label{fig:raymarching}
\end{figure}

As described in Section~\ref{sec:method}, for in-memory random access the NeuralVDB representation of choice, $[\textrm{Hash},5,4,\textrm{NN}(3)]$, combines a standard VDB tree with neural networks for the voxel values only. In Table~\ref{tab:random_queries}, we compare the performance of the in-memory random access of $[\textrm{Hash},5,4,\textrm{NN}(3)]$ and $[\textrm{Hash},5,4,3]$, implemented as NanoVDB, by randomly sampling 1M points inside the bounding box of a given volume. The NanoVDB results are generated by performing zeroth (nearest neighbor), first-order (tri-linear), and third-order (tri-cubic) interpolation using \texttt{nanovdb::SampleFromVoxels} function object for each random sample. The time-complexity of NeuralVDB is a combination of that of NanoVDB's random access tree-traversal, which is identical for the two representations, and the neural network inference applied to a subset of the original sampling points. The NeuralVDB random access is more expensive than both nearest neighbor and tri-linear interpolation of NanoVDB, but similar to third-order interpolation, and cheaper than pure neural network predictions since it prunes out queries that fall into tiles, i.e., non-voxels. 

\ken{As an additional benchmark test we implemented a simple ray-marcher that operates on $[\textrm{Hash},5,4,\textrm{NN}(3)]$, see Figure~\ref{fig:raymarching}. \doyubmr{Rendering of the  bunny model with $[\textrm{Hash},5,4,3]$ using the zeroth-order sampler took 75 ms, first-order sampler took 97 ms, and the third-order sampler took 1660 ms, compared to 1316 ms for the $[\textrm{Hash},5,4,\textrm{NN}(3)]$ grid.} 
This benchmark test illustrates that while NeuralVDB can replace OpenVDB for run-time applications like rendering that require in-memory random access, it does come with a performance trade off} \doyub{which is comparable to the higher-order samplers.} All results were measured with a single NVIDIA A40 GPU.

\subsection{Random Sampling Error}

We already showed the quantitative measurement of the NeuralVDB's reconstruction accuracy in Table~\ref{tab:non-temporal-results} and \ref{tab:temporal-results}, and the qualitative visualization in Figure~\ref{fig:sdf-error} and \ref{fig:vol-error}. Here, we show a further experiment where we compare the sampling errors between conventional grid-based interpolation methods and NeuralVDB consuming the same amount of memory. We first create a NanoVDB grid initialized with a simplex noise function. We also generate a NeuralVDB grid with approximately the same ``in-memory'' footprint, which is trained with the same noise function. We then generate 1M random sampling points and perform zeroth, first-order, and third-order queries to the NanoVDB grid and the voxel value regression for the NeuralVDB grid. We measure RMSE error for each sampling strategy to evaluate their accuracy compared to the ground truth noise function. The results are shown in Table~\ref{tab:sampling_error}. We can observe that the accuracy goes up when higher-order methods are used, and NeuralVDB can have better performance than even the third-order cubic sampling result.

\begin{table*}[t]
\centering
\resizebox{\textwidth}{!}{
\input{tables/non_temporal_results.tex}
}
\caption{List of input grid statistics for SDF models and density volumes: OpenVDB file sizes for both raw 32-bit precision with no compression and 16-bit precision with Blosc compression~\cite{blosc} in MB, NeuralVDB file size with 16-bit precision with Blosc compression in MB, number of total parameters (both learnable and static) of neural networks, number of false positive patches for the classifiers, compression ratio comparing 16-bit compressed file sizes, and evaluation metrics including IoU and mCD for the selected SDF volumes, and RMSE for the selected density volumes.}
\label{tab:non-temporal-results}
\end{table*}

\begin{table*}[t]
\centering
\resizebox{\textwidth}{!}{
\input{tables/temporal_results.tex}
}
\caption{List of input grid statistics for animated SDF models and density volumes: OpenVDB file sizes for both raw 32-bit precision with no compression and 16-bit precision with Blosc compression~\cite{blosc} in MB, NeuralVDB file size with 16-bit precision with Blosc compression in MB, number of total parameters (both learnable and static) of neural networks, number of false positive patches for the classifiers, compression ratio comparing 16-bit compressed file sizes, and evaluation metrics including IoU and mCD for the selected SDF volumes, and RMSE for the selected density volumes.}
\label{tab:temporal-results}
\end{table*}

\begin{table*}[t]
\centering
\resizebox{\textwidth}{!}{
\input{tables/hyperparameters.tex}
}
\caption{List of hyperparameters used in all the experiments, including subdomain size (in voxel dimension for a cubic subdomain), the number of layers and neurons per layer for the level-1 classifier (L-1 Net.), the tile value regressor, and the level-0 classifier (L-0 Net.), and the voxel value regressor. The activation function is either $\sin$ or ReLU, and if $\sin$ is used, the frequency parameters are noted. All these examples were trained using FFM, and the mapping scale and feature size are shown as well. Finally, learning rate (LR), LR decay rate and its interval, resampling interval, and maximum epochs for each example are listed. For the animation examples (LeVeque's Test, Smoke Plume, Ship Breach, Dust Impact, and Tornado), two different learning rates are shown where the first value is the initial (cold-start) learning rate whereas the second value is for the refinement (warm-start).}
\label{tab:hyperparameters}
\end{table*}

\begin{table}
\centering
\resizebox{\columnwidth}{!}{
\input{tables/performance.tex}
}
\caption{Encoding/decoding performance measured using multiple GPUs for the static volumes. The timing in seconds and relative scaling factor is presented for each volume.}
\label{tab:non-temporal-performance}
\end{table}

\begin{table}
\centering
\resizebox{\columnwidth}{!}{
\input{tables/random_queries.tex}
}
\caption{Random access performance measured for NanoVDB (zeroth, first, and third-order interpolation), NeuralVDB ($[\textrm{Hash},5,4,\textrm{NN}(3)]$), and pure neural networks (same structure as the voxel value regressor of the NeuralVDB) in milliseconds. For each static test model, 1M random samples with batch size of $2^{16}$ were generated within the model's bounding box.}
\label{tab:random_queries}
\end{table}

\begin{table}
\centering
\resizebox{\columnwidth}{!}{
\input{tables/accuracy.tex}
}
\caption{RMSE measured for both NanoVDB and NeuralVDB ($[\textrm{Hash},5,4,\textrm{NN}(3)]$) where both grids encode a fractal Brownian motion field~\cite{vivo2015book}. For NanoVDB, four different sampling methods are tested (zeroth, first, and third-order interpolation). Both NanoVDB and NeuralVDB have similar ``in-memory'' footprint. For each test model, 1M random samples with batch size of $2^{16}$ were generated within the model's bounding box.}
\label{tab:sampling_error}
\end{table}

\subsection{Comparison}
\label{sec:comparison}

\doyub{The goal of this paper is to effectively encode volumetric data with good reconstruction quality. Therefore, we designed our comparison experiments to focus on how well a given method can reconstruct volumes with low-quality loss for the same model sizes. We compared NeuralVDB with three different neural representation methods, including Neural Geometric Level of Details (NGLOD)~\cite{takikawa2021neural}, Variable Bitrate Neural Fields (VBNF)~\cite{takikawa2022variable}, and Instant Neural Graphics Primitives (INGP)~\cite{mueller2022instant}, as they provide compact neural representations using dedicated data structures (octree for NGLOD and VBNF or hash grid for INGP) as well as quantization (VBNF). We used Kaolin Wisp as a reference implementation for these three methods~\cite{KaolinWispLibrary}.}

\doyub{For the encoding process, the input was a mesh, and the output was a trained SDF neural model. In the case of NeuralVDB, the input mesh was converted into a narrow-band level set using OpenVDB's \texttt{vdb\_tool}. Other methods used Kaolin Wisp's mesh sampler, which utilizes an octree data structure for generating samples. While the CPU-based mesh sampler is available as part of the open source repository, we also acquired a private GPU implementation of the mesh sampler from the authors of the library. We included both performance results from the public and private codes in our comparison. For the decoding (reconstruction) process, the input was the trained model, and the output was a volume represented in OpenVDB format. For non-NeuralVDB methods, we densely sampled the bounding boxes and extracted a narrow band of the SDF volume to reconstruct VDB grids.}

\begin{figure*}[t]
\centering
\includegraphics[width=\textwidth]{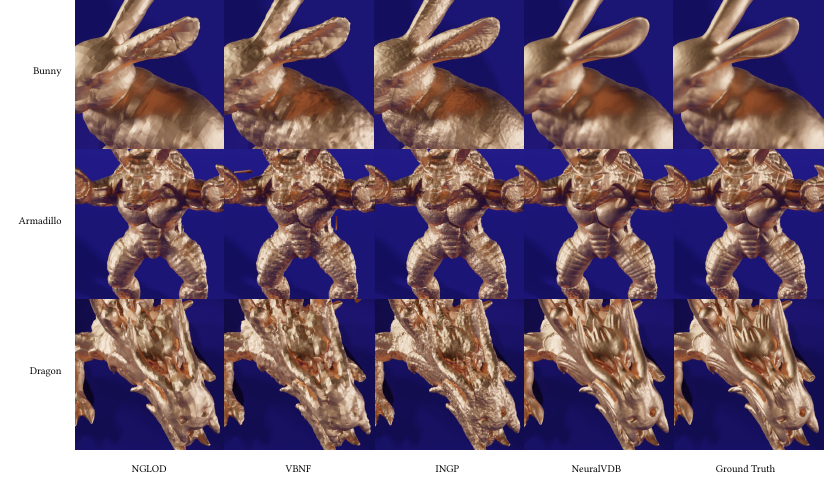}
\caption{Visualization of different neural representation methods on various SDF geometries.}
\label{fig:comparison}
\end{figure*}

\begin{table*}[t]
\centering
\resizebox{0.8\textwidth}{!}{
\input{tables/comparison.tex}
}
\caption{Performance comparison of different neural representation methods on various SDF geometries. All inputs were mesh geometries. The encoding timings include generating samples from input meshes. For non-NeuralVDB methods, we included both public open source version of the mesh sampler (Public) as well as the private GPU-accelerated version of the mesh sampler (Private) that we acquired from the authors. While NeuralVDB supports multi-GPU encoding and decoding, single GPU is used for all the experiments for the comparison}
\label{tab:comparison}
\end{table*}

\ken{In the first comparison experiment, we made each method produce similar model sizes to NeuralVDB for a given input mesh. We tested with three different input meshes (Bunny, Armadillo, and Dragon) and evaluated the IoU, mCD, encoding and decoding times. The reported encoding time includes the following steps: reading and processing of the input mesh, generation of samples, training of the model, and compression/serialization of the model to disk. Similarly, the decoding times measure \minjae{deserialization} of the model, inference, and writing back to the VDB data structure. The results are summarized in Table~\ref{tab:comparison} and visualized in Figure~\ref{fig:comparison}. NeuralVDB achieved the best performance with respect to most of the metrics, both in terms of quality and encoding/decoding timings.
A notable exception is the encoding time of the Dragon model where NGLOD with private GPU mesh sampler code was the fastest. Among the non-NeuralVDB methods, INGP achieved the best accuracy and decoding performance, since this method was specifically designed for fast inference with a hash grid that can utilize larger feature dimensions for better reconstruction quality.}

\begin{figure}
\centering
\includegraphics[width=\linewidth]{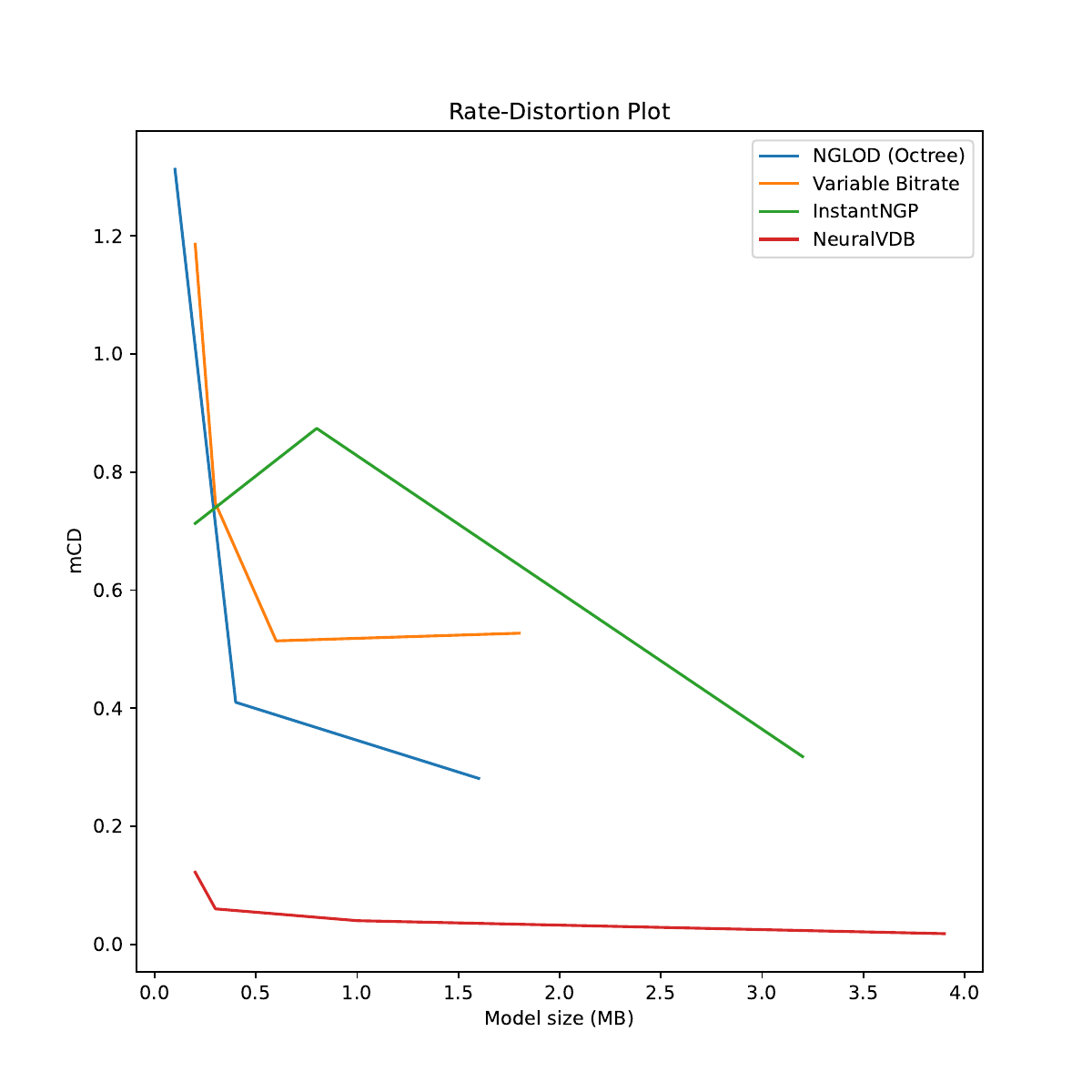}
\caption{Rate-distortion plot for different neural representation methods.}
\label{fig:rd-plot}
\end{figure}

\ken{In the second comparison experiment, we compared the rate distortion plot, which measures the distortion loss for different compression levels. We used mCD for the distortion loss and the model size of the compression level. We used the Bunny model as the input for each method. As shown in Figure~\ref{fig:rd-plot}, the results were consistent compared to the first experiment above, where NeuralVDB showed better accuracy (lower mCD) across different compression levels. Among the other methods, NGLOD performed better than other non-NeuralVDB methods as it can effectively leverage sparsity of the volume distribution. The INGP does show better accuracy over other methods for the smallest model size, and it converges slower than NGLOD with more model parameters. The VBNF also performed worse than NGLOD, which is expected as it has been found perform better on NeRF representations but exhibits high-frequency errors on SDF models~\cite{takikawa2022variable}.} 

\doyubmr{Note that the comparisons in these experiments were conducted solely on SDF \kenmr{representations}. We couldn't directly compare density volume encodings with the existing methods, as they \kenmr{only} support SDF or NeRF models. However, these methods also address sparsity using their own approaches, such as octrees or hash grids, in contrast to the VDB tree in NeuralVDB. Nonetheless, NeuralVDB has demonstrated superior performance, \kenmr{although its advantage in terms of} sparsity diminishes in denser volumes like clouds, compared to truncated SDFs. For these denser volumes, all methods would need to increase their capacity, either by expanding the MLP network to be wider and deeper or by increasing the feature vector dimension. \minjaemr{In the case of INGP, the size of the hash table is also crucial.} Therefore, we maintain that there will likely be a performance gap between NeuralVDB and other methods.}

%% file: tables/numbers.tex
\newcommand{\EMUCOMPRATIO}{40.9 }
\newcommand{\EMUHIGHRESCOMPRATIO}{11.4 }
\newcommand{\CRAWLERCOMPRATIO}{13.8 }
\newcommand{\CRAWLERHIGHRESCOMPRATIO}{3.8 }

%% file: tables/non_temporal_results.tex
\begin{tabular}{rrrrrrrrrrrr}
\toprule
        Name & Num. Active Voxels &                     Effective Res. &  VDB Raw &  VDB Comp. &  [\textrm{Hash},5,\textrm{NN}(4),\textrm{NN}(3)] & Num. Params & Num. Patches & \textbf{Comp. Ratio} &    IoU &    mCD &   RMSE \\
\midrule
       Bunny &          5,513,993 &      628 $\times$ 621 $\times$ 489 &     33.3 &       15.2 &                      0.2 &     125,379 &            0 &        \textbf{61.2} &  0.999 &  0.072 &      - \\
   Armadillo &         22,734,512 &   1276 $\times$ 1519 $\times$ 1160 &    137.7 &       63.5 &                      1.5 &     752,274 &        9,402 &        \textbf{41.3} &  0.998 &  0.115 &      - \\
      Dragon &         23,347,893 &    2023 $\times$ 911 $\times$ 1347 &    140.0 &       65.0 &                      1.8 &     889,868 &        9,172 &        \textbf{36.2} &  0.997 &  0.125 &      - \\
        Lucy &         61,305,123 &   1866 $\times$ 1073 $\times$ 3200 &    679.7 &      167.5 &                      3.3 &   1,184,774 &      134,360 &        \textbf{50.1} &  0.998 &  0.138 &      - \\
         EMU &         96,956,688 &   1481 $\times$ 2609 $\times$ 1843 &    541.8 &      232.3 &                      5.7 &   2,661,894 &       71,793 &        \textbf{40.9} &  0.999 &  0.106 &      - \\
 Thai Statue &        141,166,655 &   2358 $\times$ 3966 $\times$ 2038 &   1522.8 &      377.5 &                     13.6 &   3,812,364 &      759,320 &        \textbf{27.8} &  0.997 &  0.249 &      - \\
       Space &        165,909,193 & 32844 $\times$ 24702 $\times$ 9156 &    950.2 &      439.7 &                     14.3 &   5,995,044 &      344,405 &        \textbf{30.8} &  1.000 &  0.169 &      - \\
     Crawler &        181,196,266 &    2619 $\times$ 511 $\times$ 2149 &    846.2 &      254.3 &                     18.5 &   9,160,716 &      118,464 &        \textbf{13.8} &  0.996 &  0.174 &      - \\
 Smoke Plume &         11,111,873 &      254 $\times$ 500 $\times$ 319 &     31.4 &       24.1 &                      0.9 &     459,622 &        2,616 &        \textbf{26.7} &      - &      - &  0.081 \\
 Bunny Cloud &         19,210,271 &      577 $\times$ 572 $\times$ 438 &    139.7 &       43.8 &                      0.9 &     323,014 &       41,395 &        \textbf{48.0} &      - &      - &  0.073 \\
   Chameleon &         93,994,042 &    1016 $\times$ 1012 $\times$ 700 &    445.1 &      160.2 &                      1.1 &     592,387 &           10 &       \textbf{140.9} &      - &      - &  0.025 \\
Disney Cloud &      1,487,654,107 &   1987 $\times$ 1351 $\times$ 2449 &   3947.5 &     1491.5 &                     25.0 &  11,825,176 &      293,110 &        \textbf{59.6} &      - &      - &  0.080 \\
\bottomrule
\end{tabular}

%% file: tables/temporal_results.tex
\begin{tabular}{rrrrrrrrrrrr}
\toprule
              Name & Num. Active Voxels &                          Effective Res. &  VDB Raw &  VDB Comp. &  [\textrm{Hash},5,\textrm{NN}(4),\textrm{NN}(3)] & Num. Params & Num. Patches & \textbf{Comp. Ratio} &   IoU &   mCD &  RMSE \\
\midrule
LeVeque's Test Min &          7,084,662 &         572 $\times$  547 $\times$  547 &     81.1 &       19.6 &                                              0.6 &     333,699 &            8 &           \textbf{-} & 0.954 & 0.133 &     - \\
               Max &         29,117,298 &      1351 $\times$  1155 $\times$  1155 &    325.4 &       78.5 &                                              6.4 &   3,336,990 &        2,929 &           \textbf{-} &     1 & 0.345 &     - \\
              Mean &         17,700,052 &  1053.7 $\times$  970.6 $\times$  970.6 &    201.2 &       48.5 &                                              3.3 &   1,718,383 &           91 &        \textbf{14.7} & 0.992 & 0.167 &     - \\
   Smoke Plume Min &          9,462,168 &         231 $\times$  493 $\times$  319 &     27.2 &       20.5 &                                              1.4 &     673,126 &        2,126 &           \textbf{-} &     - &     - & 0.071 \\
               Max &         11,453,882 &         272 $\times$  500 $\times$  319 &     32.0 &       24.7 &                                              1.4 &     673,126 &        4,828 &           \textbf{-} &     - &     - & 0.075 \\
              Mean &         10,658,599 &   254.0 $\times$  496.3 $\times$  319.0 &     30.0 &       23.0 &                                              1.4 &     673,126 &        3,870 &          \textbf{17} &     - &     - & 0.073 \\
       Tornado Min &          7,084,010 &           321 $\times$ 284 $\times$ 447 &     27.1 &       16.8 &                                              0.4 &    213,350  &          819 &           \textbf{-} &     - &     - & 0.025 \\
               Max &          7,909,306 &           303 $\times$ 305 $\times$ 447 &     27.4 &       18.0 &                                              0.4 &    213,350  &        4,223 &           \textbf{-} &     - &     - & 0.035 \\
              Mean &        7,342,839.5 &     312.9 $\times$ 309.3 $\times$ 446.6 &     27.2 &       17.3 &                                              0.4 &    213,350  &        2,537 &        \textbf{40.5} &     - &     - &  0.03 \\
   Dust Impact Min &                 34 &            163 $\times$ 139 $\times$ 25 &      0.0 &        0.0 &                                              0.3 &     180,582 &            2 &           \textbf{-} &     - &     - &     0 \\
               Max &         25,553,596 &           716 $\times$ 855 $\times$ 339 &     89.2 &       55.8 &                                              2.1 &   1,083,492 &       16,872 &           \textbf{-} &     - &     - & 0.034 \\
              Mean &      13,160,681.80 &     630.4 $\times$ 720.1 $\times$ 227.3 &     46.7 &       28.5 &                                              1.5 &     771,442 &        5,105 &        \textbf{18.8} &     - &     - & 0.009 \\
   Ship Breach Min &         29,539,953 &       1295 $\times$  204 $\times$  1440 &    296.5 &       76.2 &                                              4.0 &   2,056,716 &        8,496 &           \textbf{-} & 0.989 & 0.095 &     - \\
               Max &         54,216,738 &      1728 $\times$  1419 $\times$  1970 &    596.6 &      145.2 &                                             12.1 &   6,170,148 &       96,707 &           \textbf{-} & 0.998 & 0.265 &     - \\
              Mean &         41,325,814 & 1490.7 $\times$  727.3 $\times$  1793.7 &    488.4 &      112.7 &                                              6.1 &   2,844,612 &       26,584 &        \textbf{18.4} & 0.995 & 0.131 &     - \\
\bottomrule
\end{tabular}

%% file: tables/hyperparameters.tex
\begin{tabular}{rrrrrrrrrrrrr}
\toprule
               &  Subdomain Size &     L-1 Net. & Tile Val. Net. &     L-0 Net. & Voxel Val. Net. & Activation/Freq. & FFM Scale/Size & Learning Rate & LR Decay/Interval &  Max. Epochs &  Sample Interval & Batch Size \\
\midrule
         Bunny &            1024 &  3$\times$48 &              - &  3$\times$96 &     3$\times$96 &     $\sin$ / 3.0 &        5.0/192 &         0.001 &         0.975/100 &         2500 &                1 &   $2^{16}$ \\
     Armadillo &            1024 &  3$\times$48 &              - &  3$\times$96 &     3$\times$96 &     $\sin$ / 3.0 &        5.0/192 &         0.001 &         0.975/100 &         2500 &                1 &   $2^{16}$ \\
        Dragon &            1024 &  3$\times$64 &              - & 3$\times$128 &    3$\times$128 &     $\sin$ / 1.5 &       10.0/256 &         0.001 &         0.975/100 &         2500 &                1 &   $2^{16}$ \\
          Lucy &            2048 & 3$\times$128 &              - & 3$\times$256 &    3$\times$256 &     $\sin$ / 1.5 &       10.0/256 &         0.001 &         0.975/100 &         2500 &                1 &   $2^{16}$ \\
           EMU &            2048 & 3$\times$192 &              - & 3$\times$384 &    3$\times$384 &             ReLU &       10.0/384 &         0.001 &         0.75/1000 &        10000 &              500 &   $2^{12}$ \\
   Thai Statue &            2048 & 3$\times$128 &              - & 4$\times$256 &    3$\times$256 &     $\sin$ / 1.5 &       10.0/512 &         0.001 &         0.975/100 &         2500 &                1 &   $2^{16}$ \\
         Space &            2048 &  3$\times$96 &              - & 3$\times$192 &    3$\times$192 &             ReLU &       10.0/384 &         0.001 &         0.975/100 &         2500 &                1 &   $2^{16}$ \\
       Crawler &            1536 & 3$\times$192 &              - & 4$\times$384 &    4$\times$384 &             ReLU &       20.0/768 &        0.0002 &         0.75/1000 &         6000 &              100 &   $2^{16}$ \\
   Bunny Cloud &            1024 &  3$\times$64 &    3$\times$16 & 3$\times$192 &    3$\times$192 &     $\sin$ / 3.0 &        5.0/192 &         0.001 &         0.975/100 &         2500 &                1 &   $2^{16}$ \\
     Chameleon &            1024 & 3$\times$128 &              - & 3$\times$256 &    3$\times$256 &     $\sin$ / 3.0 &       10.0/256 &         0.001 &         0.975/100 &         2500 &                1 &   $2^{16}$ \\
  Disney Cloud &            1536 & 3$\times$256 &   3$\times$128 & 4$\times$512 &    4$\times$512 &     $\sin$ / 2.0 &       20.0/512 &         0.001 &         0.75/1000 &        10000 &              500 &   $2^{12}$ \\
LeVeque's Test &            1024 &  3$\times$96 &              - & 3$\times$192 &    3$\times$192 &     $\sin$ / 1.5 &        2.0/192 &  0.001/0.0002 &         0.975/100 &         2500 &                1 &   $2^{16}$ \\
   Smoke Plume &             512 &  3$\times$48 &    3$\times$16 & 3$\times$256 &    3$\times$256 &     $\sin$ / 3.0 &       10.0/384 &  0.001/0.0001 &         0.975/100 &         2500 &                1 &   $2^{16}$ \\
   Dust Impact &             512 &  3$\times$48 &    3$\times$16 & 3$\times$128 &    3$\times$128 &     $\sin$ / 1.5 &       15.0/192 &         0.001 &         0.975/100 &         2500 &                1 &   $2^{16}$ \\
       Tornado &             512 &  3$\times$48 &    3$\times$16 & 3$\times$128 &    3$\times$128 &     $\sin$ / 1.5 &       15.0/256 &         0.001 &         0.975/100 &         2500 &                1 &   $2^{16}$ \\
   Ship Breach &            1024 &  3$\times$96 &              - & 3$\times$192 &    3$\times$256 &     $\sin$ / 1.5 &       10.0/384 &   0.001/0.001 &         0.975/100 &         5000 &                1 &   $2^{16}$ \\
\bottomrule
\end{tabular}

%% file: tables/performance.tex
\begin{tabular}{rrrrrrr}
\toprule
             & \multicolumn{3}{r}{Encoding Time} & \multicolumn{3}{r}{Decoding Time} \\
     \# GPUs &              1 &              2 &              4 &              1 &              2 &              4 \\
\midrule
       Bunny &         32.020 &              - &              - &          1.683 &              - &              - \\
             & \textbf{1.000} &              - &              - & \textbf{1.000} &              - &              - \\
\midrule
   Armadillo &         84.262 &         46.123 &         44.632 &          6.897 &              - &              - \\
             & \textbf{1.000} & \textbf{1.827} & \textbf{1.888} & \textbf{1.000} &              - &              - \\
\midrule
      Dragon &         66.244 &         38.506 &         33.369 &          7.913 &              - &              - \\
             & \textbf{1.000} & \textbf{1.720} & \textbf{1.985} & \textbf{1.000} &              - &              - \\
\midrule
        Lucy &         85.650 &         58.407 &              - &         25.530 &         17.036 &              - \\
             & \textbf{1.000} & \textbf{1.466} &              - & \textbf{1.000} & \textbf{1.499} &              - \\
\midrule
         EMU &        151.618 &         91.497 &              - &         43.943 &         30.367 &         25.176 \\
             & \textbf{1.000} & \textbf{1.657} &              - & \textbf{1.000} & \textbf{1.447} & \textbf{1.745} \\
\midrule
 Thai Statue &        148.361 &        104.713 &         99.304 &         75.917 &         51.958 &         24.852 \\
             & \textbf{1.000} & \textbf{1.417} & \textbf{1.494} & \textbf{1.000} & \textbf{1.461} & \textbf{3.055} \\
\midrule
       Space &        158.421 &        101.308 &         75.100 &         79.580 &         51.995 &         41.334 \\
             & \textbf{1.000} & \textbf{1.564} & \textbf{2.109} & \textbf{1.000} & \textbf{1.531} & \textbf{1.925} \\
\midrule
     Crawler &        284.042 &        156.179 &        110.086 &        103.198 &         59.391 &         44.362 \\
             & \textbf{1.000} & \textbf{1.819} & \textbf{2.580} & \textbf{1.000} & \textbf{1.738} & \textbf{2.326} \\
\midrule
 Bunny Cloud &         55.614 &              - &              - &          5.181 &              - &              - \\
             & \textbf{1.000} &              - &              - & \textbf{1.000} &              - &              - \\
\midrule
   Chameleon &         74.923 &              - &              - &         22.678 &              - &              - \\
             & \textbf{1.000} &              - &              - & \textbf{1.000} &              - &              - \\
\midrule
Disney Cloud &        709.446 &        431.065 &        285.415 &        397.376 &        269.628 &        180.088 \\
             & \textbf{1.000} & \textbf{1.646} & \textbf{2.486} & \textbf{1.000} & \textbf{1.474} & \textbf{2.207} \\
\bottomrule
\end{tabular}

%% file: tables/random_queries.tex
\begin{tabular}{rrrrrr}
\toprule
        Name &  NanoVDB (0) &  NanoVDB (1) &  NanoVDB (3) &  NeuralVDB &  Neural Net \\
\midrule
       Bunny &        0.107 &        0.287 &        4.548 &      2.762 &      40.481 \\
   Armadillo &        0.073 &        0.169 &        3.916 &      3.634 &      44.320 \\
      Dragon &        0.068 &        0.166 &        3.850 &      6.174 &      50.245 \\
        Lucy &        0.072 &        0.135 &        3.513 &      1.313 &      79.199 \\
         EMU &        0.090 &        0.282 &        4.068 &      6.506 &      94.817 \\
 Thai Statue &        0.073 &        0.178 &        3.897 &      6.740 &      95.810 \\
       Space &        0.058 &        0.155 &        3.763 &      5.797 &      49.250 \\
     Crawler &        0.159 &        0.968 &        5.034 &     10.345 &     156.831 \\
 Bunny Cloud &        0.074 &        0.217 &        4.579 &     11.325 &      59.108 \\
   Chameleon &        0.086 &        0.241 &        3.239 &      9.653 &      71.046 \\
Disney Cloud &        0.122 &        0.533 &        4.397 &     24.617 &     191.327 \\
\bottomrule
\end{tabular}

%% file: tables/accuracy.tex
\begin{tabular}{rrrrrrrrrr}
\toprule
Method &  NanoVDB (0) &  NanoVDB (1) &  NanoVDB (3) &  NeuralVDB \\
\midrule
  RMSE &        0.206 &        0.157 &        0.149 &      0.133 \\
\bottomrule
\end{tabular}

%% file: tables/comparison.tex
\begin{tabular}{llrrrrr}
                   &                    & \multicolumn{1}{l}{Model File Size (MB)} & \multicolumn{1}{l}{IoU} & \multicolumn{1}{l}{mCD}      & \multicolumn{1}{l}{Encoding (Public/Private) (sec.)} & \multicolumn{1}{l}{Decoding (sec.)} \\ \hline
Bunny              & NGLOD              & 0.2                                      & 0.966                   & 0.516                        & 96.318 / 99.078                     & 8.815                               \\
34,835 vertices    & VBNF               & 0.2                                      & 0.980                   & 0.762                        & 182.459 / 163.218                   & 24.722                              \\
                   & INGP               & 0.2                                      & 0.992                   & 0.449                        & 630.898 / 342.754                   & 8.063                               \\
                   & \textbf{NeuralVDB} & 0.2                                      & \textbf{0.997}          & \textbf{0.122}               & \textbf{62.048}                     & \textbf{1.683}                      \\ \hline
                   &                    & \multicolumn{1}{l}{}                     & \multicolumn{1}{l}{}    & \multicolumn{1}{l}{}         & \multicolumn{1}{l}{}                & \multicolumn{1}{l}{}                \\ \hline
Armadillo          & NGLOD              & 1.8                                      & 0.984                   & 0.853                        & 193.397 / 119.767                   & 82.909                              \\
172,976 vertices   & VBNF               & 1.7                                      & 0.941                   & 1.084                        & 1365.301 / 1055.914                 & 1065.290                            \\
                   & INGP               & 1.8                                      & 0.989                   & 0.767                        & 1690.559 / 358.348                  & 47.917                              \\
                   & \textbf{NeuralVDB} & 1.5                                      & \textbf{0.998}          & \textbf{0.115}               & \textbf{88.558}                     & \textbf{6.897}                      \\ \hline
                   & \textbf{}          & \textbf{}                                & \multicolumn{1}{l}{}    & \multicolumn{1}{l}{}         & \multicolumn{1}{l}{}                & \multicolumn{1}{l}{}                \\ \hline
Dragon             & NGLOD              & 1.9                                      & 0.773                   & {\color[HTML]{333333} 1.032} & 2121.700 / \textbf{157.234}         & 140.994                             \\
5,832,139 vertices & VBNF               & 1.8                                      & 0.929                   & 1.313                        & 9203.716 / 1133.834                 & 1205.431                            \\
                   & INGP               & 1.8                                      & 0.969                   & 0.784                        & 45833.001 / 435.927                 & 70.087                              \\
                   & \textbf{NeuralVDB} & 1.8                                      & \textbf{0.997}          & \textbf{0.125}               & 191.716                             & \textbf{7.913}                      \\ \hline
\end{tabular}

%% file: 5_discussion.tex
In this paper, we presented NeuralVDB, a new, highly compact VDB framework using hierarchical neural networks. We combined the effectiveness of the standard sparse VDB structure and the highly efficient compression capability of neural networks. To further leverage the high compression ratio of neural networks, we use them to encode \emph{both} voxel values as well as the topology (i.e., node and tile connectivity) of the two lowest levels of the tree structure itself. This results in a novel representation, dubbed $[\textrm{Hash},5,\textrm{NN}(4),\textrm{NN}(3)]$, that reduces the memory footprint of the already compact VDB, with up to a factor of 100 in some cases. We also propose a NeuralVDB configuration, denoted $[\textrm{Hash},5,4,\textrm{NN}(3)]$, which balances memory reduction and random access performance. While both configurations feature highly attractive characteristics in terms of the reduced memory footprints, they are by no means silver bullets. More to the point, we are not proposing that NeuralVDB can replace standard VDBs for all applications. In fact, we primarily recommend $[\textrm{Hash},5,\textrm{NN}(4),\textrm{NN}(3)]$ as a very efficient but lossy offline representation.

As indicated already there are some limitations to NeuralVDB that we seek to improve in future work. While NeuralVDB can encode and decode most of the examples in a couple of minutes, some examples like the Disney Cloud takes nearly five minutes to encode and three minutes to decode. Also, the random query performance is comparable to the third-order interpolation of NanoVDB, but still slower than the first-order sampler, which is typically used in computer graphics applications. We expect to achieve improved performance by further reducing the size of the neural networks, e.g.,~by means of improved feature mapping like neural hash encoding~\cite{mueller2022instant} and/or applying mixed-precision inference. Specifically for encoding/training, data-driven approaches like MetaSDF~\cite{sitzmann2020metasdf} can help warm starting the training process. Such warm starting feature has already been leveraged in our animated examples with great success. Also, while most of the offline compressors like Zlib~\cite{gailly2004zlib} or Blosc~\cite{blosc} have a few control parameters, NeuralVDB has even more hyperparameters that need to be specified for optimal performance. This usability issue can be improved by systematic/automated parameter selection, potentially using data-driven approaches. \doyubmr{Additionally, in the context of the temporal encoder, although initializing the network with the previous frame significantly diminishes artifacts, there is still a noticeable level of reconstruction artifacts present.} Lastly, NeuralVDB shares one fundamental limitation with NanoVDB, notably not shared with the standard VDB, namely that it assumes the tree and its values to be fixed. This is an assumption that we also plan to relax in future work.

%% file: 8_acknowledgements.tex
We thank Nvidia for supporting this project and in particular Christopher Horvath, Alexandre Sirois-Vigneux, Greg Klar, Jonathan Leaf, Andre Pradhana, and Wil Braithwaite for the water simulation and rendering of Ship Breach, and Nuttapong Chentanez, Matthew Cong, Stefan Jeschke, Eric Shi, Ed Quigley, and Byungsoo Kim for proofreading our paper. We also thank to Towaki Takikawa, Or Perel, and Clement Fuji Tsang for their help on conducting the comparison experiment using Kaolin Wisp\cite{KaolinWispLibrary}.

%% file: 9_appendix.tex
\section{Random Access in VDB}
\label{appx:random-access-vdb}
Random (i.e., coordinate-based) access to values in a VDB structure is fast (on average constant time) due to a unique caching mechanism and the fact that the tree structure has a fixed depth of only four levels. Whenever a random value query is performed, a value accessor caches all the nodes visited. For subsequent queries, the cached nodes are initially visited bottom-up, and the first node that contains the new query point is used as the starting point for a top-down traversal, which also updates the cache with newly visited nodes. Consequently, a value accessor effectively performs a bottom-up, versus a traditional top-down, tree traversal, which is very fast for typical access patterns, like Finite-Difference stencils that are spatially coherent.

\section{NanoVDB}
\label{appx:nanovdb}
\ken{The open source C++ implementation of VDB, dubbed OpenVDB, makes use of several libraries that only work on CPUs, or more to the point not on GPUs. NanoVDB~\cite{museth2021nanovdb} addressed this limitation by offering C++ and C99 implementations of the VDB tree structure without any external library dependencies. Consequently, NanoVDB runs on both CPUs and GPUs, and supports most graphics APIs including CUDA, DX12, OptiX, OpenGL, OpenCL, Vulcan and GLSL. However, one limitation is that NanoVDB assumes the topology of the tree to be static, which follows from the entire tree can be serialized (or linearized) into a single continuous block of memory. Other than GPU support, NanoVDB offers another advantage over OpenVDB, namely in-memory compression by means of variable bit-rate quantization with dithering to randomize the inevitable quantization noise. This typically reduces the memory footprints of NanoVDB volumes by a factor of 4-6 relative to OpenVDB representations, at the cost of small quantization errors and the assumption of fixed trees, which is ideal for especially rendering and some simulation applications.}

\section{Effect of Training with Sparsity Information}
\label{appx:effect-of-training-with-sparsity}

\begin{figure*}[t]
\centering
\includegraphics[width=\textwidth]{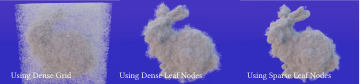}
\caption{\ken{Comparison of the effects that sparse vs dense representations have on training of sparse volumes}. \doyub{The left image is trained with a dense grid with no sparsity information. The middle image is trained with a sparse blocked grid, but without active voxel masks. The right image is trained with \ken{a VDB grid that offers both sparse nodes and active voxel masks}. All experiments were trained with the same network architecture and hyper-parameters.}}
\label{fig:active-voxels}
\end{figure*}

\ken{As an ablation study, we performed an experiment where the Bunny Cloud model is trained on 1) a dense grid, 2) a block grid (represented by dense leaf nodes in a VDB tree), and 3) sparse voxels represented by the active voxel in a VDB grid. We trained these models with identical configurations including the MLP architecture as well as training parameters. As shown in Figure~\ref{fig:active-voxels}, the results from the dense grid contain the highest amount of noise, whereas the other models show much better visual quality. The noise is still visible in the blocked (2) fog volume, which does not make use of the active voxel masks in VDB. However, when using the sparse VDB voxel representation (3), the same network can effectively reconstruct the model with low noise.}

\section{Effect of Activation Functions}
\label{appx:activation}

\begin{figure*}[t]
\centering
\includegraphics[width=\textwidth]{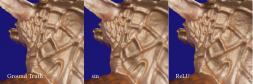}
\caption{\doyub{Comparison of the effect of $\sin$ (middle column) and ReLU activation functions (right-most column) for the Armadillo example. The ground truth model (left) has both smooth geometry as well as flat surfaces from the coarse input mesh. The $sin$ activation function tends to smooth out the sharp edges while ReLU can introduce more high-frequency noise.}}
\label{fig:activations}
\end{figure*}

While the ReLU activation function works for most of the cases, we noticed that different activation functions could affect the reconstruction quality as well as the convergence rate. In our experiments, the ReLU works great for flat, structured, or artificial models, while the $\sin$ and $\tanh$ function can work better for smooth or unstructured models as shown in Figure~\ref{fig:activations}. Furthermore, although it is redundant to the Fourier feature mapping, we noticed that the $\sin$ function, as shown in SIREN~\cite{sitzmann2020implicit} also can be used for specific smooth and unstructured models and also accelerate the convergence compared to ReLU or $\tanh$ functions.

\section{Heuristic Estimation of Hyper-parameters}
\label{appx:hyperparameters}

Table~\ref{tab:hyperparameters} lists several hyperparameters for NeuralVDB that impact both accuracy and efficiency. In this section, we elaborate on the effect of each hyperparameter and provide heuristics for determining their values.

The subdomain size affects both training accuracy and time. If too large, most of the subdomains will be empty, wasting computing resources. If the domain is too small, the cost of dispatching query points can be non-negligible, and overlapping halo regions can become dominant, which in turn results in redundant computation. We picked a subdomain size for each experiment with a multiple of 512, which proved sufficient to efficiently subdivide the examples studies in this paper.

For the network parameters, level-1 networks use half of the width of the level-0 networks. We found that for most examples, three layers were sufficient for the desired tolerances, while four layers are used for volumes with more details to capture. As mentioned in \ref{appx:activation}, we used either $\sin$ or ReLU activations depending on how smooth or structured the input volume is. The frequency of $\sin$ activation ranged from 1.5 to 3.0. 

The parameters for Fourier feature mapping are determined by the width of the network. For a wide network, which normally means there are high-frequency details to capture, the same number of mapped features (FFM size) to the network width and a larger FFM scale are used (see Section~\ref{sec:feature-mapping}).

The sampling strategy for the batches was either drawing $2^{16}$ random samples for each epoch or resampling a subset of input voxels for a given interval and drawing smaller batches ($2^{12}$) for each epoch. In the latter case, the number of a subset to be resampled is determined by the number of resampling intervals (either 100 or 500 in our examples) times batch size. This resampling is used when fine details with thin structures are critical.

For the network optimizer, a learning rate of 0.001 was used for most of the experiments, except the Crawler model which has uniquely complex geometric features. We decayed the learning rate with 0.975 with an interval of 100 when no subset resampling was used. When we performed the resampling, we used a decay rate of 0.75 with an interval of 1000. The number of maximum epochs was 2500 for non-resampled cases. More epochs were used for the resampled cases.

In general, for less artificially shaped geometry or small volumes, a level-0 network with width 128, depth 3, $\sin$ activation with frequency 1.5, matching feature mapping size with the width, and FMM scale of 5 or greater proved to be a good starting point. (Level-1 network should be half of the width.) Similar to our examples, a learning rate of 0.001, decay rate and interval of 0.975 and 100, and maximum epochs of 2500 with a batch size of $2^{16}$ should be sufficient for most cases. For a structured geometry or a volume with high-frequency details, wider networks with the subset resampling approach and its parameter set from one of our examples should be a good baseline.